%% file: acl_latex.tex
\pdfoutput=1

\documentclass[11pt]{article}
\usepackage[dvipsnames]{xcolor}
\usepackage{pgfplots}
\usepackage{multicol}
\usepackage{verbatim}
\usepackage{amsmath}

\usepackage[preprint]{acl}

\usepackage{times}
\usepackage{latexsym}

\usepackage[T1]{fontenc}

\usepackage[utf8]{inputenc}

\usepackage{microtype}

\usepackage{inconsolata}

\usepackage{graphicx}
\usepackage{tabularx}
\usepackage{multirow}
\usepackage{amsfonts}
\usepackage{amssymb}
\usepackage{xspace}
\usepackage{lipsum}
\usepackage{tcolorbox}
\usepackage{adjustbox}
\usepackage{booktabs}
\usepackage{makecell}
\usepackage{url}
\usepackage{enumitem}
\setlist[enumerate]{topsep=0pt, partopsep=0pt, parsep=0pt, itemsep=0pt}
\setlist[itemize]{topsep=0pt, partopsep=0pt, parsep=0pt, itemsep=0pt}
\usepackage{listings}
\usepackage{xcolor}
\usepackage{inconsolata}
\usepackage{caption}
\usepackage{float}
\usepackage{pgfplots}
\usepackage{subcaption}
\pgfplotsset{compat=1.17}
\usetikzlibrary{positioning,fit,arrows.meta,calc,backgrounds}
\usepackage{pgfplotstable}
\usepackage{amsthm}

\newcommand{\drexel}{%
  \hspace{1pt}
  \begingroup\normalfont
  \includegraphics[height=1.3\fontcharht\font`\B]{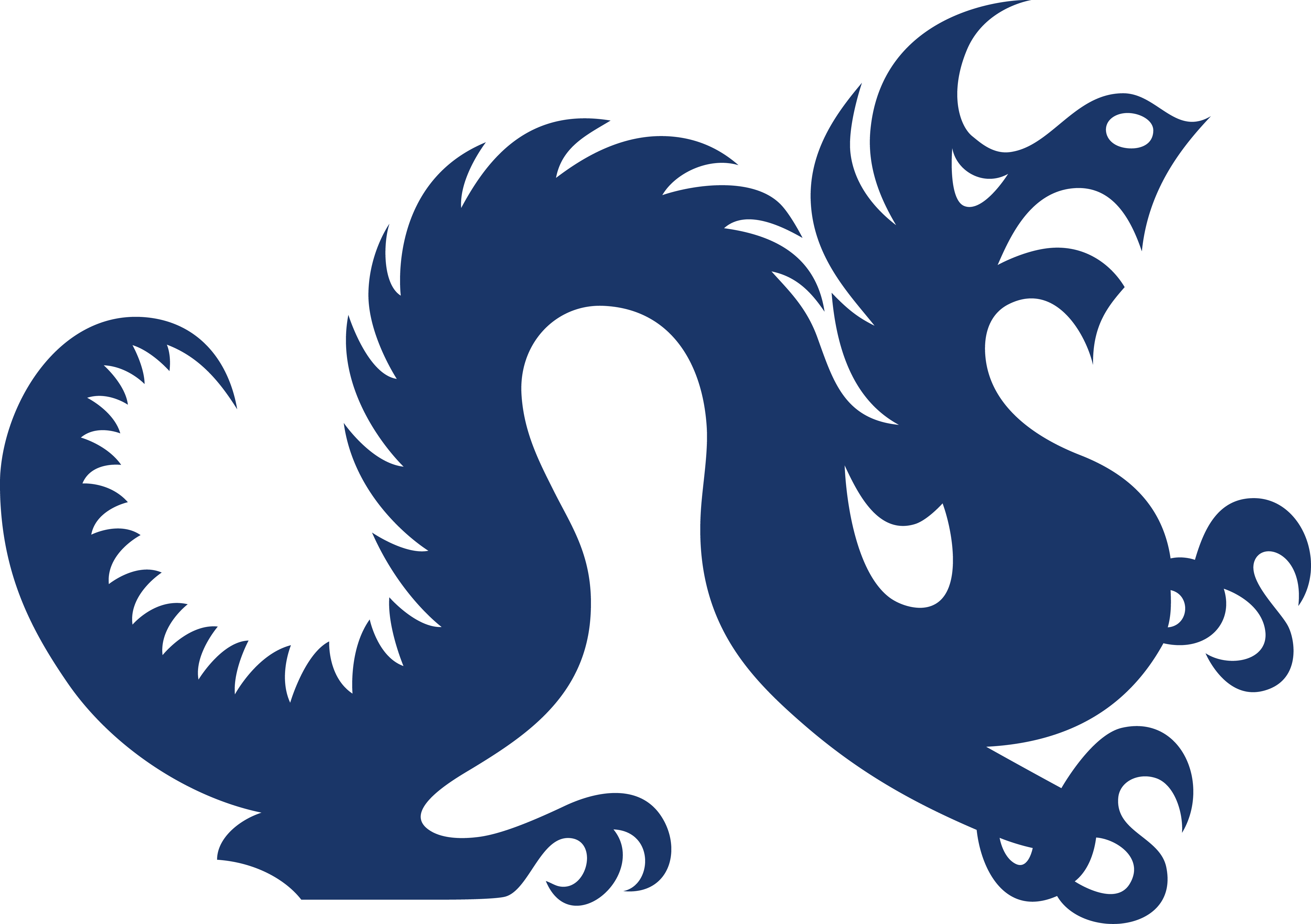}%
  \endgroup
  \hspace{1pt}
}
\newcommand{\aitwo}{%
  \hspace{1pt}%
  \begingroup\normalfont
  \includegraphics[height=1.3\fontcharht\font`\B]{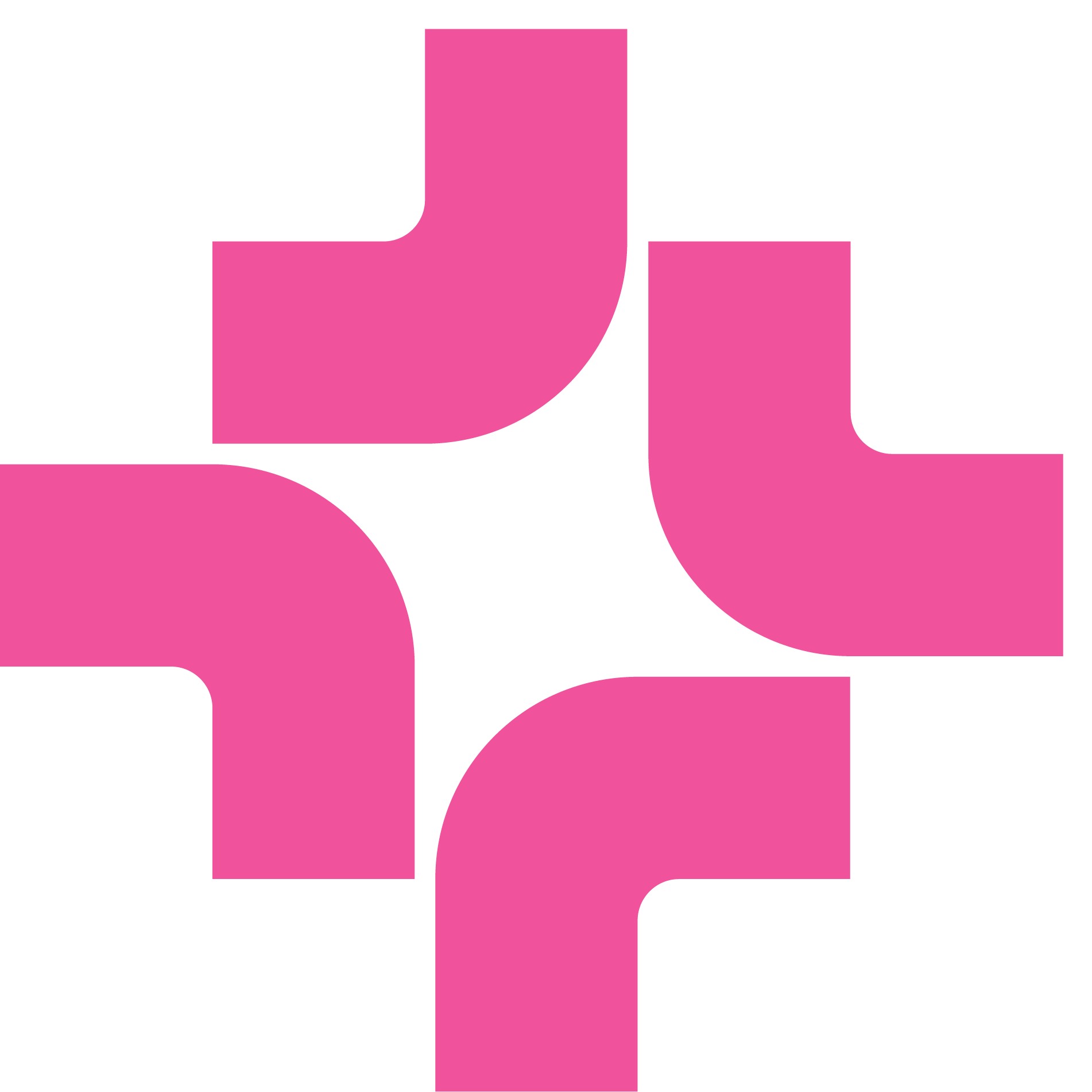}%
  \endgroup
  \hspace{1pt}%
}

\pgfplotsset{
  taskbar/.style 2 args={
    width=\linewidth, height=4.5cm,
    ybar, bar width=8pt, bar shift=0pt,
    enlarge x limits=0.25, ymin=0, ymax=#1,
    xtick={1,2,3}, xticklabels={Base,CautA,CautB},
    tick label style={font=\footnotesize},
    label style={font=\footnotesize},
    title style={font=\small},
    ylabel={#2}, xlabel={Prompt},
    every axis plot/.append style={draw=none},
    legend style={font=\scriptsize, at={(0.5,-0.25)}, anchor=north, legend columns=3},
  }
}

\newfloat{listing}{tp}{lol}
\floatname{listing}{Listing}
\floatstyle{plain}
\restylefloat{listing}

\usepackage{tcolorbox}
\tcbuselibrary{listings, skins, breakable}
\definecolor{spuriousred}{RGB}{180,30,30}
\definecolor{genuinegreen}{RGB}{30,120,60}
\definecolor{boxgray}{RGB}{245,245,245}
\definecolor{spuriousbg}{RGB}{255,245,245}
\definecolor{genuinebg}{RGB}{245,255,248}
\definecolor{codeyellow}{RGB}{255,252,220}

\definecolor{cQ3}{RGB}{0,114,178}       % blue
\definecolor{cOSS}{RGB}{230,159,0}      % orange
\definecolor{cQ25}{RGB}{0,158,115}      % teal green
\definecolor{cDS}{RGB}{204,121,167}     % mauve pink

\lstdefinestyle{mystyle}{
    basicstyle=\ttfamily\small,
    breaklines=true,
    frame=single,
    backgroundcolor=\color{gray!5},
    keywordstyle=\color{blue!80!black}\bfseries,
    commentstyle=\color{gray!70}\itshape,
    stringstyle=\color{orange!80!black},
    showstringspaces=false,
    tabsize=2,
    rulecolor=\color{black},
    framesep=6pt,
    columns=fullflexible,
    language=Python,
    morekeywords={True,False,None,def,return,if,else,for,in,range,print,json},
}

\lstdefinestyle{jsonstyle}{
    basicstyle=\ttfamily\small,
    breaklines=true,
    frame=single,
    backgroundcolor=\color{gray!5},
    stringstyle=\color{green!60!black},
    numberstyle=\color{blue!70!black},
    keywordstyle=\color{purple!80!black}\bfseries,
    showstringspaces=false,
    tabsize=2,
    rulecolor=\color{black},
    framesep=6pt,
    columns=fullflexible,
}
\lstdefinestyle{prompt}{style=mystyle, language={}, frame=none, backgroundcolor={}, keywordstyle={}, stringstyle={}, commentstyle={}, basicstyle=\ttfamily\scriptsize, literate={—}{{\textemdash}}1}

\title{Solver-Like Behavior in Large Language Models as Formalizers}

\author{Rikhil Amonkar \drexel \enspace Stuti Mohan \drexel \enspace Kevin Nguyen \drexel \\ \textbf{Ceyhun Efe Kayan} \drexel \enspace \textbf{Ronan Le Bras} \aitwo \enspace \textbf{Li Zhang} \drexel \\
  \drexel Drexel University \enspace \aitwo Allen Institute for AI \hspace{4pt} \\
  {\texttt{\{rikhil.amonkar|harry.zhang}@drexel.edu\}}
}

\begin{document}
\maketitle

\begin{abstract}

In the LLM-as-formalizer paradigm, a large language model translates a problem into a formal program and leaves the search to a solver, gaining the solver's guarantees. Formalizers are now often reasoning models that write a long reasoning trace before the program, yet they are judged only by the program's output. We find that the trace routinely does the solver's job, assigning, testing and ruling out values until it often reaches the full answer; we call this \textit{solver-like behavior} (SLB). We define an ontology of SLB over the trace, the program and its output, and use LLM judges, checked against human labels, to label 2{,}000 generations from three reasoning models on four constraint satisfaction datasets, each formalized in a declarative and an imperative formal language. SLB appears in 85\% of traces and takes 24\% of reasoning tokens, yet the program usually recomputes the answer, so the effort is redundant. The 9\% of programs that inherit the trace's answer also inherit its errors, and pass output-based evaluation whenever that answer is correct. The amount of SLB is not associated with accuracy, and SLB is more common in imperative Python than in declarative Z3 or PDDL. Cautionary prompts reduce it by only a quarter, but hiding the problem's values behind placeholders, which the program receives only at run time, removes 97\% of it without loss of accuracy. Formalizers should therefore be evaluated by how their programs are reached, not only by what they output.\footnote{Our code is available at \url{https://github.com/rikhil-amonkar/solver-like-behavior-in-llms}.}

\end{abstract}

\section{Introduction}
\label{sec:intro}

In the LLM-as-formalizer paradigm \cite{tantakoun-etal-2025-llms}, a large language model (LLM) does not solve a problem but translates it into a formal program, such as a planning domain or a set of constraints, which a solver executes \cite{huang-zhang-2025-limit, lin2025zebralogic, amonkar2025naturallanguageplanningcoding}. The LLM handles language, and the solver performs the search with guarantees that the LLM's own reasoning lacks \cite{wei-etal-2025-plangenllms}. Formalizers today are often reasoning models that write a long reasoning trace before the program. Yet they are judged almost only by whether the program's output is correct, which cannot tell a program that searches from one that restates an answer the model already worked out in its trace. The trace itself has gone unexamined: work on reasoning traces, which finds them often unfaithful or wasteful \cite{lanham2023measuringfaithfulnesschainofthoughtreasoning, arcuschin2025chain, chen2025do, sui2025stopoverthinking}, studies models that solve the problem themselves, and formalizer traces have been discussed only in passing \cite{amonkar2025naturallanguageplanningcoding}.

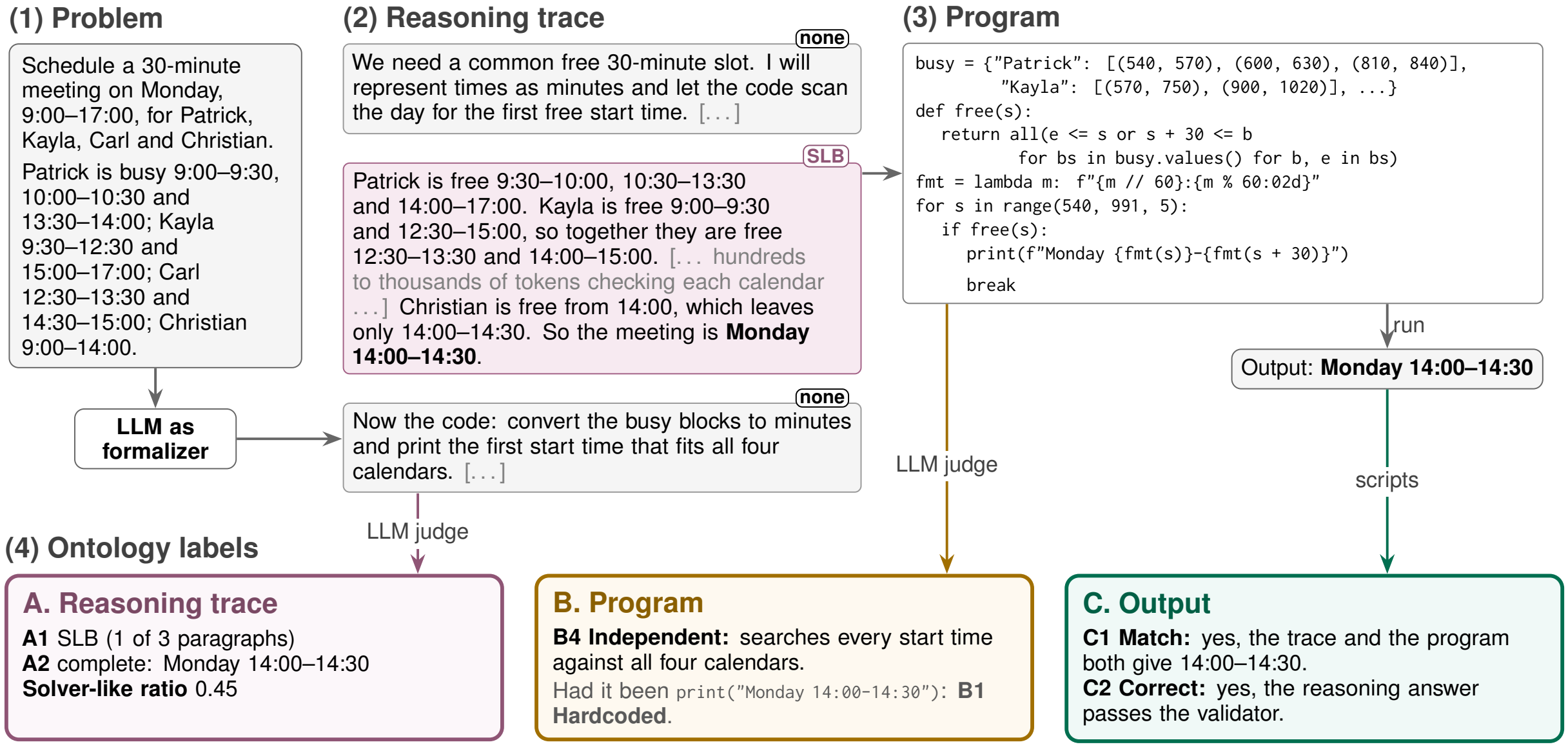
\begin{figure*}[t!]
\centering
\resizebox{\textwidth}{!}{%
\begin{tikzpicture}[
  font=\sffamily\small,
  >={Stealth[length=2.2mm]},
  every node/.append style={execute at begin node={\hyphenpenalty=10000 \exhyphenpenalty=10000 }},
  stp/.style={font=\sffamily\bfseries\footnotesize, text=black!75, anchor=south west, inner sep=0pt},
  para/.style={draw=#1, rounded corners=2pt, align=left, text width=5.6cm, inner sep=3pt,
               font=\sffamily\scriptsize},
  pill/.style={draw=#1!70!black, fill=white, rounded corners=2pt, inner sep=1.2pt,
               font=\sffamily\tiny\bfseries, text=#1!70!black},
  panel/.style={draw=#1!70!black, very thick, rounded corners=4pt, fill=#1!6, inner sep=5pt},
  head/.style={font=\sffamily\bfseries\footnotesize, text=#1!60!black, anchor=north west, inner sep=0pt},
  lbl/.style={align=left, text width=5.2cm, inner sep=0pt, font=\sffamily\scriptsize, anchor=north west},
  lab/.style={font=\sffamily\scriptsize, text=black!75, fill=white, inner sep=1.5pt},
]
\newcommand{\cdl}[1]{{\ttfamily\fontsize{6.3}{7.6}\selectfont #1}}

\node[draw=black!60, rounded corners=3pt, fill=black!4, align=left, text width=3.0cm, inner sep=4pt,
      font=\sffamily\scriptsize, anchor=north west] (prob) at (0,0)
  {Schedule a 30-minute meeting on Monday, 9:00--17:00, for Patrick, Kayla, Carl and Christian.\\[2pt]
   Patrick is busy 9:00--9:30, 10:00--10:30 and 13:30--14:00; Kayla 9:30--12:30 and 15:00--17:00; Carl 12:30--13:30 and 14:30--15:00; Christian 9:00--14:00.};
\node[stp] at ($(prob.north west)+(0,3pt)$) {(1) Problem};
\node[draw=black!60, rounded corners=3pt, fill=white, align=center, font=\sffamily\scriptsize\bfseries,
      text width=1.6cm, inner sep=3pt, anchor=north] (llm) at ($(prob.south)+(0,-0.45)$) {LLM as formalizer};
\draw[->, thick, black!60] (prob.south) -- (llm.north);

\newcommand{\elide}[1]{\textcolor{black!50}{[\ldots\ #1\ \ldots]}}
\node[para=black!40, fill=black!4, anchor=north west] (p1) at (3.75,0)
  {We need a common free 30-minute slot. I will represent times as minutes and let the code scan the day for the first free start time. \textcolor{black!50}{[\ldots]}};
\node[para=cDS!80!black, fill=cDS!16, below=9pt of p1] (p2)
  {Patrick is free 9:30--10:00, 10:30--13:30 and 14:00--17:00. Kayla is free 9:00--9:30 and 12:30--15:00, so together they are free 12:30--13:30 and 14:00--15:00.
   \elide{hundreds to thousands of tokens checking each calendar}
   Christian is free from 14:00, which leaves only 14:00--14:30. So the meeting is \textbf{Monday 14:00--14:30}.};
\node[para=black!40, fill=black!4, below=9pt of p2] (p3)
  {Now the code: convert the busy blocks to minutes and print the first start time that fits all four calendars. \textcolor{black!50}{[\ldots]}};
\foreach \p/\c/\t in {p1/black/none, p2/cDS/SLB, p3/black/none}
  \node[pill=\c, anchor=south east] at ($(\p.north east)+(-4pt,-1.6pt)$) {\t};
\node[fit=(p1)(p3), inner sep=0pt] (R) {};
\node[stp] at ($(p1.north west)+(0,3pt)$) {(2) Reasoning trace};
\draw[->, thick, black!60] (llm.east) -- (llm.east-|R.west);

\node[draw=black!50, rounded corners=2pt, fill=white, align=left, inner sep=4pt, anchor=north west,
      text width=6.9cm] (prog) at ($(R.north east)+(0.45,0)$)
  {\cdl{busy = \{"Patrick": [(540, 570), (600, 630), (810, 840)],\\
   \hspace*{4.3em}"Kayla": [(570, 750), (900, 1020)], ...\}\\
   def free(s):\\
   \hspace*{1.3em}return all(e <= s or s + 30 <= b\\
   \hspace*{5.2em}for bs in busy.values() for b, e in bs)\\
   fmt = lambda m: f"\{m // 60\}:\{m \% 60:02d\}"\\
   for s in range(540, 991, 5):\\
   \hspace*{1.3em}if free(s):\\
   \hspace*{2.6em}print(f"Monday \{fmt(s)\}-\{fmt(s + 30)\}")\\
   \hspace*{2.6em}break}};
\node[stp] at ($(prog.north west)+(0,3pt)$) {(3) Program};
\node[draw=black!60, rounded corners=3pt, fill=black!4, font=\sffamily\scriptsize, inner sep=3pt,
      anchor=north east] (out) at ($(prog.south east)+(0,-0.5)$) {Output: \textbf{Monday 14:00--14:30}};
\draw[->, thick, black!60] (out.north|-prog.south) -- (out.north) node[midway, right, lab, fill=none] {run};
\draw[->, thick, black!60] (R.east|-prog.west) -- (prog.west);

\coordinate (lt) at ($(prob.west|-R.south)+(0,-0.95)$);
\node[head=cDS] (ah) at ($(lt)+(0.15,-0.15)$) {A. Reasoning trace};
\node[lbl] (a) at ($(ah.south west)+(0,-3pt)$)
  {\textbf{A1} SLB (1 of 3 paragraphs)\\
   \textbf{A2} complete: Monday 14:00--14:30\\
   \textbf{Solver-like ratio} 0.45};
\node[head=cOSS] (bh) at ($(ah.north west)+(5.95,0)$) {B. Program};
\node[lbl] (b) at ($(bh.south west)+(0,-3pt)$)
  {\textbf{B4 Independent:} searches every start time against all four calendars.\\[1pt]
   {\color{black!65}Had it been \cdl{print("Monday 14:00-14:30")}: \textbf{B1 Hardcoded}.}};
\node[head=cQ25] (chd) at ($(bh.north west)+(5.95,0)$) {C. Output};
\node[lbl] (c) at ($(chd.south west)+(0,-3pt)$)
  {\textbf{C1 Match:} yes, the trace and the program both give 14:00--14:30.\\
   \textbf{C2 Correct:} yes, the reasoning answer passes the validator.};
\begin{scope}[on background layer]
\node[panel=cDS, fit=(ah)(a)(a.south|-b.south)(a.east)] (A) {};
\node[panel=cOSS, fit=(bh)(b)] (B) {};
\node[panel=cQ25, fit=(chd)(c)(c.south|-b.south)] (C) {};
\end{scope}
\node[stp] at ($(A.north west)+(0,3pt)$) {(4) Ontology labels};

\coordinate (ja) at ($(R.south west)+(0.85,0)$);
\coordinate (jb) at ($(prog.south west)+(0.5,0)$);
\draw[->, thick, cDS!70!black] (ja) -- (ja|-A.north) node[midway, lab] {LLM judge};
\draw[->, thick, cOSS!70!black] (jb) -- (jb|-B.north) node[pos=0.6, lab] {LLM judge};
\draw[->, thick, cQ25!70!black] (out.south) -- (out.south|-C.north) node[midway, lab] {scripts};
\end{tikzpicture}}
\caption{Overview on a calendar scheduling problem (1). An LLM as formalizer writes a reasoning trace (2) and then a program (3), which is run and scored. The middle paragraph is \textcolor{cDS!80!black}{\textbf{solver-like}}: it solves the problem in the trace, finding first the common free intervals and then the meeting time, which is the solver's job; in real traces such a passage runs to hundreds or thousands of tokens. Our ontology (\S\ref{sec:ontology}) labels every generation (4): whether the reasoning trace is solver-like (A), how much of the reasoning answer the program inherits (B), and whether the program's output matches the reasoning answer, which is checked for correctness (C). Here the program searches independently, so the solver-like behavior, 45\% of the trace, was wasted.}
\label{fig:eval_pipeline}
\end{figure*}

We find that formalizers routinely do the solver's job themselves, a phenomenon we call \emph{solver-like behavior} (SLB): before writing the program, the reasoning trace assigns, tests and rules out values of the problem's variables, often until it reaches the full answer (Figure~\ref{fig:eval_pipeline}). This defeats the purpose of formalization in two ways. It is costly: constraint satisfaction is NP-complete \cite{10.1145/800157.805047}, and solving in tokens grows with the instance: on Meeting Planning, solver-like tokens nearly triple from the smallest to the largest problems, though the program usually searches anyway. And it can silently void the solver's guarantee: a program that prints the answer from the trace, i.e.\ \emph{inherits} it, passes output-based evaluation while also inheriting the trace's errors, so where outputs are not independently validated, a formalization that looks correct may be no more verifiable than chain-of-thought.

We study SLB in three reasoning models on four constraint satisfaction datasets, each formalized in a declarative formal language (Z3 or PDDL) and an imperative one (Python). We define an ontology over the trace, the program and its output (Figure~\ref{fig:ontology}), label 2{,}000 generations with LLM judges that agree with blind human labels on 94\% of paragraphs, and test four hypotheses (\S\ref{sec:results}). \textbf{(H1) SLB is common and redundant:} 85\% of traces contain it, yet the program usually recomputes the answer. \textbf{(H2) Its forms differ in harm:} the 9\% of programs that inherit the reasoning answer also inherit its errors, and pass output-based evaluation whenever that answer is correct; hardcoding is rare and mostly comes from one model on one dataset. \textbf{(H3) Its amount is not associated with accuracy} on the same problem. \textbf{(H4) The formal language matters most:} Python draws more SLB than Z3 or PDDL in every dataset. Cautionary prompts cut SLB by about a quarter, while hiding the problem's values behind placeholders that the program receives at run time removes 97\% of it without loss of accuracy (\S\ref{sec:deter}). SLB is thus pervasive and mostly redundant, and occasionally voids the solver's guarantee.

\section{Related Work}
\label{sec:related}

\textbf{LLMs as formalizers.} An LLM paired with an external solver writes PDDL for classical planners \cite{liu2023llm+, guan2023leveraging, xie2023translating, zuo-etal-2025-planetarium, hu2025text2worldbenchmarkinglargelanguage, huang-zhang-2025-limit, gong2025zeroshotiterativeformalizationplanning, huang2026languagemodelplannerformalizer}, logic for theorem provers and SAT/SMT solvers \cite{pan-etal-2023-logic, olausson-etal-2023-linc, ye2023satlm}, optimization models \cite{li2026constructingindustrialscaleoptimizationmodeling} and scheduling constraints \cite{amonkar2025naturallanguageplanningcoding}; \citet{tantakoun-etal-2025-llms} survey the planning line, and program-aided reasoning \cite{gao2023pal, chen2023program, lyu-etal-2023-faithful} likewise delegates computation to an interpreter. These works evaluate the program, by whether it executes, matches a reference or yields a valid solution; we examine the reasoning that precedes it.

\textbf{Reasoning traces.} Chains of thought often misrepresent the computation behind the answer: models omit cues that sway them \cite{turpin2023language, chen2025reasoningmodelsdontsay}, rationalize after the fact \cite{arcuschin2025chain}, and keep their answer when the reasoning is truncated or corrupted \cite{lanham2023measuringfaithfulnesschainofthoughtreasoning}. Other work dissects traces into behaviors such as verification and backtracking \cite{marjanovic2025thoughtology, gandhi2025cognitivebehaviors} or finds the sentences that steer the outcome \cite{bogdan2025thoughtanchors}. Reasoning models also spend far more tokens than a problem needs \cite{chen2025do, sui2025stopoverthinking, cuadron2025dangeroverthinking}, which motivates budgets on reasoning length \cite{muennighoff2025s1} and measures of effort beyond length \cite{chen2026thinkdeepjustlong}, while accuracy can still collapse as problems grow complex \cite{shojaee2025illusionthinkingunderstandingstrengths}. All of this work studies models that solve the problem themselves. For a formalizer, solver-like behavior is a specific waste: its tokens duplicate the solver's work and grow on the hard instances that formalization is meant to offload.

\textbf{Delegation and hardcoding.} LLMs often reason in text when code would serve better \cite{chen2025steering, chen2025codesteer}, or call tools they do not need \cite{qian2025smart, wang2025actingless}; solver-like behavior is a failure to delegate that persists after the model has written a program. Its extreme form, a program that prints the trace's answer, resembles reward hacking in code generation, where models hardcode test outputs \cite{baker2025monitoring, zhong2025impossiblebench}; training against a monitor of the trace can drive such behavior into less visible forms \cite{baker2025monitoring}, much as our cautionary prompts turn literal itineraries into fixed orders (\S\ref{sec:deter}).

\section{An Ontology of Solver-Like Behavior}
\label{sec:ontology}

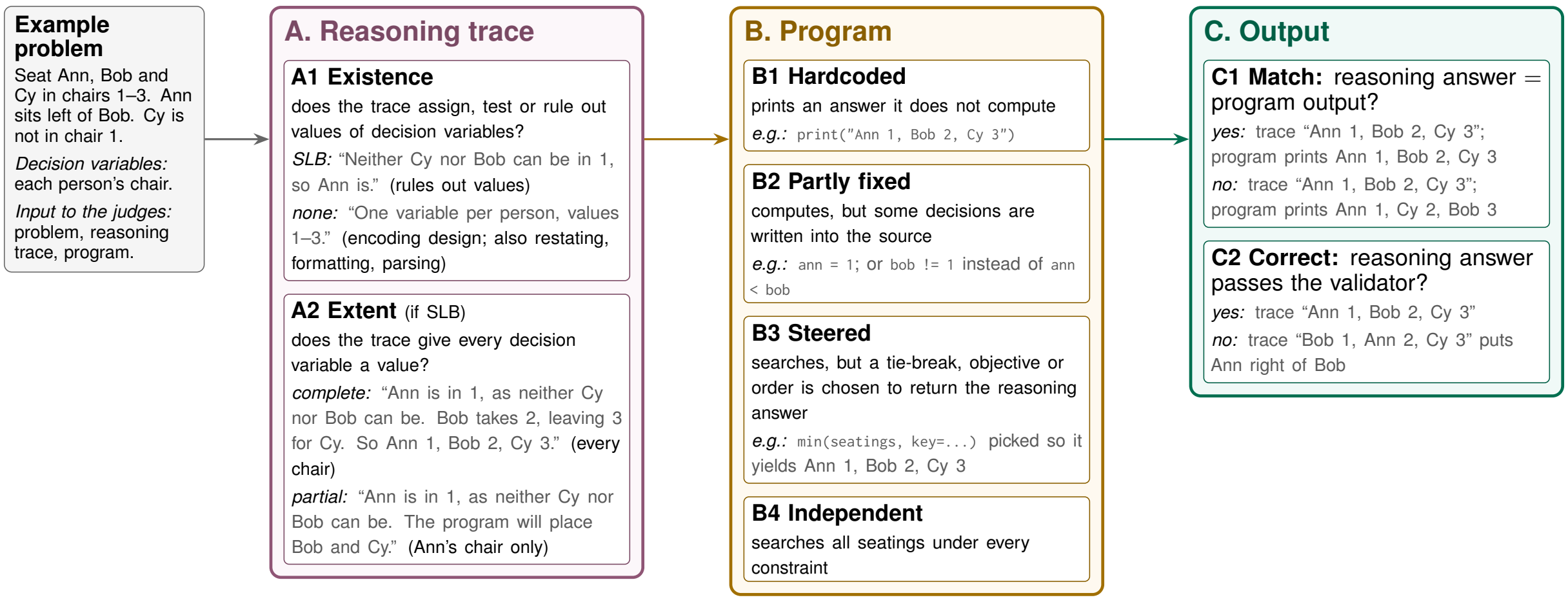
\begin{figure*}[t!]
\centering
\resizebox{\textwidth}{!}{%
\begin{tikzpicture}[
  font=\sffamily\small,
  >={Stealth[length=2.2mm]},
  every node/.append style={execute at begin node={\hyphenpenalty=10000 \exhyphenpenalty=10000 }},
  panel/.style={draw=#1!70!black, very thick, rounded corners=4pt, fill=#1!6, inner sep=5pt},
  head/.style={font=\sffamily\bfseries, text=#1!60!black, anchor=west, inner sep=0pt},
  box/.style={draw=#1!60!black, rounded corners=2pt, fill=white, align=left,
              text width=4.6cm, inner sep=3pt, font=\sffamily\footnotesize},
  inbox/.style={draw=black!60, rounded corners=3pt, fill=black!4, align=left,
                text width=2.5cm, inner sep=4pt, font=\sffamily\scriptsize},
  lab/.style={font=\sffamily\scriptsize, text=black!75, fill=white, inner sep=1.5pt},
]
\newcommand{\ex}[2]{\\[1pt]{\scriptsize\textit{#1:} \textcolor{black!65}{#2}}}
\newcommand{\q}[1]{``#1''}
\newcommand{\cd}[1]{{\ttfamily\fontsize{6.5}{7.5}\selectfont #1}}

\node[head=cDS] (ah) at (0,0) {A. Reasoning trace};
\node[box=cDS, anchor=north west] (a1) at ($(ah.south west)+(0,-5pt)$)
  {\textbf{A1 Existence}\\[1pt] {\scriptsize does the trace assign, test or rule out values of decision variables?}
   \ex{SLB}{\q{Neither Cy nor Bob can be in 1, so Ann is.}} {\scriptsize(rules out values)}
   \ex{none}{\q{One variable per person, values 1--3.}} {\scriptsize(encoding design; also restating, formatting, parsing)}};
\node[box=cDS, below=5pt of a1] (a2)
  {\textbf{A2 Extent} {\scriptsize(if SLB)}\\[1pt] {\scriptsize does the trace give every decision variable a value?}
   \ex{complete}{\q{Ann is in 1, as neither Cy nor Bob can be. Bob takes 2, leaving 3 for Cy. So Ann 1, Bob 2, Cy 3.}} {\scriptsize(every chair)}
   \ex{partial}{\q{Ann is in 1, as neither Cy nor Bob can be. The program will place Bob and Cy.}} {\scriptsize(Ann's chair only)}};
\begin{scope}[on background layer]
\node[panel=cDS, fit=(ah)(a1)(a2)] (A) {};
\end{scope}

\node[head=cOSS] (bh) at (6.4,0) {B. Program};
\node[box=cOSS, anchor=north west] (b1) at ($(bh.south west)+(0,-5pt)$)
  {\textbf{B1 Hardcoded}\\[1pt] {\scriptsize prints an answer it does not compute}
   \ex{e.g.}{\cd{print("Ann 1, Bob 2, Cy 3")}}};
\node[box=cOSS, below=5pt of b1] (b2)
  {\textbf{B2 Partly fixed}\\[1pt] {\scriptsize computes, but some decisions are written into the source}
   \ex{e.g.}{\cd{ann = 1}; or \cd{bob != 1} instead of \cd{ann < bob}}};
\node[box=cOSS, below=5pt of b2] (b3)
  {\textbf{B3 Steered}\\[1pt] {\scriptsize searches, but a tie-break, objective or order is chosen to return the reasoning answer}
   \ex{e.g.}{\cd{min(seatings, key=...)} picked so it yields Ann 1, Bob 2, Cy 3}};
\node[box=cOSS, below=5pt of b3] (b4)
  {\textbf{B4 Independent}\\[1pt] {\scriptsize searches all seatings under every constraint}};
\begin{scope}[on background layer]
\node[panel=cOSS, fit=(bh)(b1)(b4)] (B) {};
\end{scope}

\node[head=cQ25] (ch) at (12.8,0) {C. Output};
\node[box=cQ25, anchor=north west] (c1) at ($(ch.south west)+(0,-5pt)$)
  {\textbf{C1 Match:} reasoning answer $=$ program output?
   \ex{yes}{trace \q{Ann 1, Bob 2, Cy 3}; program prints Ann 1, Bob 2, Cy 3}
   \ex{no}{trace \q{Ann 1, Bob 2, Cy 3}; program prints Ann 1, Cy 2, Bob 3}};
\node[box=cQ25, below=5pt of c1] (c2)
  {\textbf{C2 Correct:} reasoning answer passes the validator?
   \ex{yes}{trace \q{Ann 1, Bob 2, Cy 3}}
   \ex{no}{trace \q{Bob 1, Ann 2, Cy 3} puts Ann right of Bob}};
\begin{scope}[on background layer]
\node[panel=cQ25, fit=(ch)(c1)(c2)] (C) {};
\end{scope}

\node[inbox, anchor=north east] (in) at ($(A.north west)+(-0.9,0)$)
  {{\footnotesize\textbf{Example problem}}\\[2pt]
   Seat Ann, Bob and Cy in chairs 1--3. Ann sits left of Bob. Cy is not in chair~1.\\[3pt]
   \textit{Decision variables:} each person's chair.\\[3pt]
   \textit{Input to the judges:} problem, reasoning trace, program.};

\draw[->, thick, black!60] (in.east) -- (in.east-|A.west);
\draw[->, thick, cOSS!70!black] (A.east|-in.east) -- (B.west|-in.east);
\draw[->, thick, cQ25!70!black] (B.east|-in.east) -- (C.west|-in.east);
\end{tikzpicture}}
\caption{Ontology of solver-like behavior, with an example of each node on a toy seating problem (left). \textbf{A} describes the reasoning trace: whether it contains solver-like behavior (A1), and how much of an answer it reaches (A2). \textbf{B} describes how much of the reasoning answer the program inherits rather than computes; B1--B3 require A1 = SLB. \textbf{C} compares the program's output with a complete reasoning answer (C1), and checks the reasoning answer with the validator (C2).}
\label{fig:ontology}
\end{figure*}

We view each task as a constraint satisfaction problem: a set of \emph{decision variables}, the unknowns that together make up an answer (e.g., when a meeting starts, where each object goes, or which action comes next in a plan), and constraints on their values. In the LLM-as-formalizer paradigm, encoding the variables and constraints is the model's job, and finding values that satisfy them is the solver's. Figure~\ref{fig:ontology} illustrates every definition below on a toy problem: seat Ann, Bob and Cy in chairs 1--3, with Ann left of Bob and Cy not in chair~1. Its decision variables are each person's chair.

\textbf{Solver-like behavior} (SLB) is text in the reasoning trace that does the solver's job on the instance: assigning, testing or ruling out values of its decision variables. Designing the encoding, restating the problem's data or constraints, choosing the output format, parsing, and interpreting an ambiguous constraint are not solver-like, unless the reading is chosen by solving under each one. A formalizer's \emph{generation} has two parts: a \emph{reasoning trace} (\emph{trace} for short), the natural-language text the model writes first, and a \emph{program}, whose \emph{output} is what it prints when run. An answer stated in the trace is the \emph{reasoning answer}. We describe solver-like behavior along these three components: the trace (A), the program (B), and its output (C).

\textbf{A. Reasoning trace} is read from the trace alone.
\textbf{A1 (existence)} is \emph{SLB} if the trace assigns, tests or rules out values of decision variables, and \emph{none} otherwise. Arguments about the solution set that commit no values, such as uniqueness, are rare and not counted.
\textbf{A2 (extent)} applies when A1 is \emph{SLB}. It is \emph{complete} if every decision variable receives a value and \emph{partial} otherwise, including a tentative answer the trace later abandons.
The \textbf{solver-like ratio} is the share of the trace that is solver-like.

\textbf{B. Program} records how much of the reasoning answer the program inherits rather than computes. A program \emph{inherits} a decision when it takes the value from the trace instead of computing it; B1--B3 are the forms this takes.
\textbf{B1 Hardcoded:} it prints an answer it does not compute, so the formalization adds no guarantee.
\textbf{B2 Partly fixed:} it computes the answer, but decisions made in the trace are written into its source. This shrinks the solver's search, which can be harmless or even necessary, as when a numerical BlocksWorld constraint, which PDDL cannot express, must be turned into fluents in the trace.
\textbf{B3 Steered:} it searches, but its tie-break, objective or iteration order is chosen so that the search returns the reasoning answer.
\textbf{B4 Independent:} it searches under every constraint, so any SLB in the trace was wasted.
Only decisions the trace had to solve for count as inherited; a restriction read off the problem statement is part of the encoding. B1--B3 presuppose A1 = \emph{SLB}, and B does not apply when the program does not run.

\textbf{C. Output} compares the program's output with the reasoning answer, and applies when A2 is \emph{complete} and the program runs.
\textbf{C1 (match)} asks whether the reasoning answer equals the program's output.
\textbf{C2 (correct)} asks whether the reasoning answer passes the validator used to score the program.
Together they distinguish programs that merely reproduce the reasoning answer from programs that override an incorrect one or are led astray by it.

\section{Methods}

\label{sec:formulation}

\paragraph{Tasks, Datasets and Formal Languages.}

Given a natural-language constraint satisfaction problem (CSP) \cite{russell2016artificial}, the model must write a program in a formal language (FL) that returns a valid solution when executed. We sample 100 instances from each of four benchmarks. Calendar Scheduling and Meeting Planning \cite{zheng2024naturalplanbenchmarkingllms} require meeting times and multi-location itineraries that satisfy availability, duration, and travel-time constraints. ZebraLogic \cite{lin2025zebralogic} consists of logic grid puzzles. BlocksWorld \cite{ipc} is classical planning. Full task descriptions are in Appendix~\ref{app:tasks}.

We formalize every dataset in two FLs; following the formalizer literature, FL here includes general-purpose programming languages. The two differ in who searches for a solution, which is what solver-like behavior is about. A \emph{declarative} program states what a solution is and a solver finds one; an \emph{imperative} program states how to find one, so the model designs the search. Our imperative FL is Python with only the standard library. Our declarative FL is PDDL \cite{mcdermott1998pddl} for BlocksWorld, solved by a classical planner, and Z3 for the other datasets: Python programs that state constraints through the API of the Z3 SMT solver \cite{demoura2008z}, which returns a satisfying assignment. Programs that use another library, or Z3 programs that never call the solver, are flagged as not written in the requested FL (at most 5\% of any run; Table~\ref{tab:runs}).

\paragraph{Prompt.}

Each model receives one prompt per dataset and FL: a prefix that instructs it to formalize the problem in the FL and let the program solve it, rather than solving it in the reasoning trace, followed by the task's output format (Appendix~\ref{app:prompts}). All Z3 prompts give the import and coding steps of the ZebraLogic Z3 prompt \cite{lin2025zebralogic}. The PDDL prompt supplies the ground-truth problem PDDL as a reference and asks for a domain and a problem; the Python BlocksWorld prompt gives only the natural-language problem, which states the same initial and goal states.

\paragraph{Models.}

We evaluate three reasoning models that write a reasoning trace before the program. We need the complete trace: closed frontier models such as Claude, GPT-5 and Gemini return at most a summary of their reasoning through their APIs, from which solver-like behavior cannot be measured. DeepSeek-V4-Flash \cite{deepseekai2025deepseekv32pushingfrontieropen} runs via API, and GPT-OSS-120B \cite{openai2025gptoss120bgptoss20bmodel} and Qwen3.8-27B \cite{qwen3technicalreport} locally with vLLM. DeepSeek is run in one FL per dataset (Python for Calendar Scheduling and Meeting Planning, Z3 for ZebraLogic, PDDL for BlocksWorld), because the model behind its API changed between our runs (Appendix~\ref{app:models}).

\paragraph{Metrics.}

We report the ontology labels of \S\ref{sec:ontology}, the solver-like ratio, and program accuracy for every generation.

\textbf{LLM judges.} No script can tell whether a passage does the solver's job, and the traces are too many and too long to label by hand. We therefore label A1, A2 and B with LLM judges, using Gemma~4~31B-it \cite{gemma4technicalreport} (FP8), which is not one of the evaluated models. Traces are too long to judge at once, so we cut each into segments of at most 2{,}000 tokens at paragraph boundaries and give the A1 judge one segment at a time, with the problem, to label every paragraph as \emph{SLB} or \emph{none}; a trace is \emph{SLB} if any paragraph is, and the solver-like ratio is the share of the trace's characters in such paragraphs. The A2 judge extracts the reasoning answer from the \emph{SLB} paragraphs, and the B judge compares the program with the trace. C1 and C2 are scripts. Against reference labels covering both FLs of every dataset, the judges agree on 90--98\% of cases (Table~\ref{tab:reliability}). Judge designs and further checks are in Appendix~\ref{app:ontology-judges}, labeled examples in Appendix~\ref{app:examples}.

\begin{table}[t]
\centering
\small
\setlength{\tabcolsep}{2.5pt}
\begin{tabular}{llrr}
\toprule
Node & Reference & Agree & 95\% CI \\
\midrule
A1 paragraph & human, blind & 103/110 & .87--.97 \\
\quad \emph{SLB} precision & human, blind & 38/42 & .78--.96 \\
\quad \emph{SLB} recall & human, blind & 38/41 & .81--.97 \\
A2 extent & human + script & 39/40 & .87--1.00 \\
B program & model, blind & 19/20 & .76--.99 \\
B program (held out) & human, not blind & 38/40 & .83--.99 \\
C1, C2 & \multicolumn{3}{l}{deterministic scripts} \\
\bottomrule
\end{tabular}
\caption{Agreement of each ontology judge with reference labels, with Wilson 95\% confidence intervals. A1: one annotator, labeling blind (70 development and 40 held-out paragraphs). A2: \emph{complete} labels verified by humans, \emph{partial} ones by searching the trace for the program's answer and having a second model read the passages found. B: blind reference labels by a second model; held out: a human shown the judge's evidence, on programs labeled B1--B4. A1 and held-out B references cover both FLs of every dataset. C1 and C2 are computed by scripts from the A2 extraction, so they need no agreement check.}
\label{tab:reliability}
\end{table}

\textbf{Accuracy.} A program is correct if it runs and its output satisfies every constraint of the problem (Appendix~\ref{app:tasks}).

\section{Results}
\label{sec:results}

We label 2{,}000 generations: Qwen and GPT-OSS in both FLs of each dataset and DeepSeek in one, each on 100 problems under the prompt above, and test the four hypotheses of \S\ref{sec:intro}. Confidence intervals are 95\% bootstrap intervals over problems; Appendix~\ref{app:full-results} lists the labels of every run.

\begin{figure*}[t]
\centering

\includegraphics[width=\textwidth]{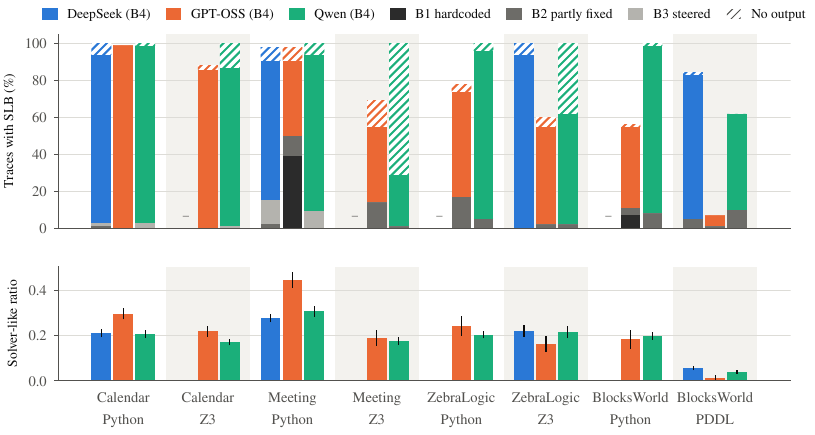}
\caption{H1 and H2: solver-like behavior is common, and the program rarely inherits it. For each dataset and FL, the share of traces with SLB (A1) and the mean solver-like ratio, with 95\% confidence intervals. Each SLB bar is split by the program's label (B): programs that inherit the reasoning answer (B1--B3, grey) at the bottom, independent programs (B4) in the model's colour, and programs with no output (missing, failed, or printed nothing) hatched; most of these are Qwen's Z3 programs, which call Z3 methods that do not exist. DeepSeek is run in one FL per dataset (--).}

\label{fig:h1}
\end{figure*}

\subsection*{H1: SLB is common and redundant}

We call solver-like behavior practically significant in a trace when it takes at least 10\% of the trace or at least 1{,}000 tokens. \textbf{By either measure, it is the norm.} 85\% of traces contain SLB (91\% outside BlocksWorld in PDDL), 71\% spend at least 10\% of their tokens on it, and 47\% at least 1{,}000 tokens (Figure~\ref{fig:h1}). Across all traces, 24\% of reasoning tokens are solver-like. A DeepSeek trace spends on average 6{,}123 of its 26{,}572 reasoning tokens on solver-like behavior, a Qwen trace 3{,}788 of 15{,}310, and a GPT-OSS trace 575 of 1{,}892. The one setting where it is rare is BlocksWorld in PDDL (51\% of traces, ratio 0.04). Stray paragraphs do not drive this: only 29 of the 1{,}699 traces with SLB spend under 2\% of their text on it, and in 77\% of all traces the SLB paragraphs alone state a complete answer.

\textbf{It grows with the problem.} From the smallest to the largest third of problems, solver-like tokens per trace rise from 3{,}873 to 10{,}971 on Meeting Planning (by number of friends) and from 532 to 4{,}148 on ZebraLogic (by grid size), as accuracy falls from 0.70 to 0.51 and from 0.90 to 0.78; GPT-OSS on ZebraLogic is the exception. On Calendar and BlocksWorld they do not grow.

\textbf{The program rarely uses it.} Of the generations whose trace contains SLB, the program computes the answer anew (B4; Figure~\ref{fig:h1}, top) in 89\% for DeepSeek and 77\% for Qwen and GPT-OSS. Yet the two usually agree: where the trace states a complete answer and the program prints one, an independent program prints exactly the reasoning answer in 88\% of cases (1{,}027 of 1{,}174; C1). The solving duplicates the work the program then does.

\subsection*{H2: SLB takes forms that differ in harm}

\textbf{SLB usually reaches a complete answer.} When a trace contains SLB, it states a complete answer in 87\% (DeepSeek), 86\% (GPT-OSS) and 96\% (Qwen) of cases (A2), i.e.\ the very answer the program is meant to compute. The reasoning answer is wrong in 2\% (DeepSeek), 11\% (Qwen) and 13\% (GPT-OSS) of the complete answers that can be checked (C2).

\textbf{Inheritance is rare and concentrated.} Of the 1{,}791 programs that ran, 163 inherit the reasoning answer (grey in Figure~\ref{fig:h1}, top, which shows the 157 whose trace contains SLB): 47 are hardcoded (B1), 40 of them GPT-OSS on Meeting Planning in Python; 88 are partly fixed (B2), mostly on BlocksWorld, Meeting Planning and ZebraLogic; and 28 are steered (B3), mostly Meeting programs in Python that try friends in the order of the reasoning answer. As B1 implies, 44 of the 45 hardcoded programs whose output can be compared print exactly the reasoning answer (C1).

\textbf{Inheritance passes on errors in the reasoning answer.} When the reasoning answer is wrong, none of the 19 hardcoded programs and 1 of the 13 partly fixed ones is correct, against 32\% of the 84 independent programs, which recompute the answer and can recover. Recomputing is no guarantee, though: 37 of those 84 independent programs print the same wrong answer as the trace (C1), so a program can share the trace's error without taking any value from it, unless the B judge misses subtler inheritance. Steering is harmless: both steered programs with a wrong reasoning answer are correct. When the reasoning answer is correct, all four kinds of program are correct at least 93\% of the time, so output-based evaluation cannot tell a hardcoded program from an independent one: all 26 hardcoded programs with a correct reasoning answer pass it.

\begin{figure}[t]
\centering
\includegraphics[width=\columnwidth]{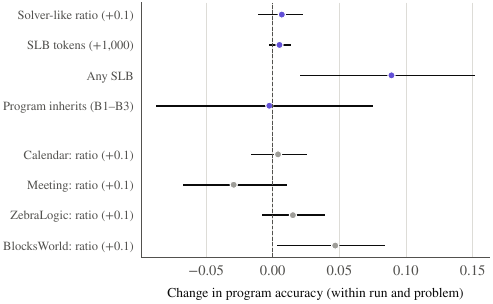}
\caption{H3: the amount of solver-like behavior is not associated with accuracy. Change in the probability that the program is correct per unit of each measure, from a linear model with fixed effects for model, dataset and FL and for the problem, with 95\% confidence intervals. Grey: the ratio's effect within each dataset.}
\label{fig:h3}
\end{figure}

\begin{figure}[t]
\centering
\includegraphics[width=\columnwidth]{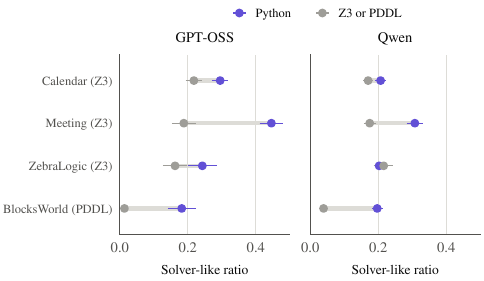}
\caption{H4: on the same 100 problems, Python draws more solver-like behavior than each dataset's declarative FL (in parentheses). Mean solver-like ratio per model, with 95\% confidence intervals.}
\label{fig:h4}
\end{figure}

\subsection*{H3: The amount of SLB is not associated with accuracy}

Harder problems may draw both more SLB and more errors, so we compare generations of the same problem: a linear probability model of program correctness with fixed effects for the problem and for each model, dataset and FL. \textbf{The amount of solving has no detectable effect} (Figure~\ref{fig:h3}). Ten more points of solver-like ratio change accuracy by $+0.007$ $[-0.011, +0.022]$, and 1{,}000 more solver-like tokens by $+0.005$ $[-0.002, +0.013]$; within datasets the ratio's effect lies between $-0.03$ and $+0.02$, except on BlocksWorld ($+0.047$ $[+0.004, +0.083]$). Programs that inherit the reasoning answer are not less accurate on average ($-0.00$ $[-0.09, +0.07]$). Traces with no SLB are less often correct ($-0.09$ $[-0.15, -0.02]$ relative to traces with some), but the gap lies in programs that fail to run: 25\% of them crash, against 12\%, and among programs that run the gap is $-0.04$ $[-0.09, +0.03]$. We cannot rule out that working through the problem helps a model write a program that runs. The interventions of \S\ref{sec:deter} agree: removing solver-like behavior does not lower accuracy.

\subsection*{H4: SLB differs by formal language}

\textbf{On the same problems, Python draws more SLB than either Z3 or PDDL, in every dataset} (Figure~\ref{fig:h4}; pooled over the two models, with Qwen on ZebraLogic the one exception). The ratio is higher in Python by $0.06$ $[0.04, 0.08]$ on Calendar, $0.20$ $[0.17, 0.23]$ on Meeting Planning and $0.03$ $[0.01, 0.06]$ on ZebraLogic than in Z3, and by $0.16$ $[0.14, 0.18]$ on BlocksWorld than in PDDL. In a linear model of the ratio with all three factors (Qwen and GPT-OSS; $R^2 = 0.27$), removing the FL loses $0.123$ of explained variance, the dataset $0.057$, and the model $0.008$. Using only the FLs DeepSeek is run in, where all three models are run, the model explains $0.011$.

\textbf{The dataset matters through the search the FL leaves to the model.} The FL effect is largest where the declarative FL takes over a search that is hard to program: in PDDL the planner finds the plan, and in Z3 the solver chooses Meeting Planning's friend order, which in Python the model must program as a search over orders. On Calendar (a trivial search) and ZebraLogic (permutations in Python) the FLs differ little. With the FL fixed, the datasets differ in Python (Meeting Planning 0.38, the others 0.19--0.25) but not in Z3 (0.18--0.19). The two prompts of a dataset also differ in more than the FL (\S\ref{sec:formulation}): the PDDL prompt supplies the reference problem file, which may account for part of the BlocksWorld gap, so the Z3 comparisons are the cleaner test of the FL.

\section{Deterring Solver-Like Behavior}
\label{sec:deter}

\begin{figure*}[t]
\centering
\includegraphics[width=\textwidth]{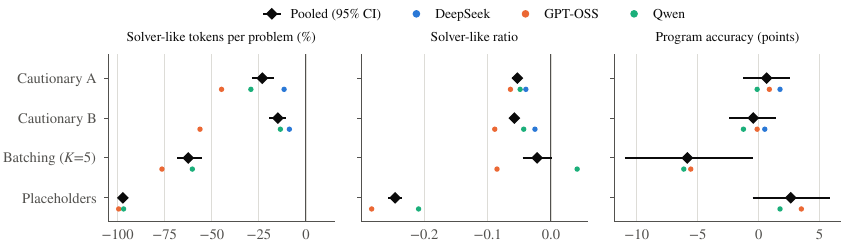}
\caption{Placeholders all but remove solver-like behavior without costing accuracy; cautionary prompts remove about a quarter of it, and batching shortens all reasoning at a cost in accuracy. Change from the baseline prompt on the same problems in solver-like tokens per problem (\% of the baseline mean), solver-like ratio, and program accuracy (points): pooled estimate with 95\% bootstrap confidence interval, and one dot per model. Each method is measured where it was run: cautionary prompts for every model, dataset and FL (2{,}000 pairs), batching ($K=5$) for Qwen and GPT-OSS in every dataset and FL except PDDL (1{,}400), placeholders for Qwen and GPT-OSS on Calendar and Meeting Planning in Python and Z3 (800). The subsets differ, so the text compares each method with the cautionary prompts on its own problems.}
\label{fig:deter}
\end{figure*}

We test three ways to deter SLB: warning against it in the prompt, batching several problems into one prompt, and hiding the problem's values behind placeholders. Each generation is compared with the baseline generation of the same problem (Figure~\ref{fig:deter}).

\subsection*{Cautionary prompts}

We add one of two warnings to the prompt of \S\ref{sec:formulation} (Appendix~\ref{app:prompts}; Listings~\ref{lst:baseline}--\ref{lst:cautb}). \textbf{Cautionary A} frames solver-like behavior by its outcome, precomputing the answer in the trace and hardcoding it, and \textbf{Cautionary B} by the act, substituting the given values into arithmetic in the trace. Each illustrates the contrast with a math word problem, so no task-specific solution leaks. We run both for every model, dataset and FL of \S\ref{sec:results}.

\textbf{Cautionary prompts reduce solver-like behavior by about a quarter, without removing it or changing accuracy} (Figure~\ref{fig:deter}). From a baseline of 0.20, the ratio falls by $0.053$ $[0.045, 0.061]$ under Cautionary A and $0.058$ $[0.051, 0.065]$ under Cautionary B, and solver-like tokens by 689 and 443 per trace (23\% and 15\%). Yet most traces still contain SLB: the share with SLB falls by 4 and 9 points, almost all of it GPT-OSS ($-10$ and $-19$ points), while DeepSeek ($+1$, $-4$) and Qwen ($-1$, $-1$) barely change. Programs that inherit the reasoning answer fall by 4 points. Accuracy does not change ($+0.01$ $[-0.01, +0.02]$ and $-0.00$ $[-0.02, +0.01]$), so reducing solver-like behavior neither helps nor hurts, as H3 found by observation.

\textbf{The prompts remove hardcoded answers more than inherited decisions.} GPT-OSS's 40 hardcoded Meeting programs in Python fall to 0 and 1, but its partly fixed ones go from 11 to 15 and 7: the program now encodes the problem, yet still follows a friend order chosen in the trace.

\subsection*{Batching and placeholders}

A trace can only assign, test or rule out values it is given. Two changes to the problem make this harder while leaving the formalization task intact. \textbf{Batching} gives $K$ problems of the same dataset in one prompt and asks for $K$ separate programs, each headed by its problem's number: the encoding can be designed once, but solving in prose must be done $K$ times. \textbf{Placeholders} replace every value of a problem by a name (\texttt{Person1}, \texttt{Day1}, \texttt{T1} for clock times, \texttt{D1} for durations, \texttt{L1} for locations, and a table \texttt{TRAVEL} for Meeting Planning's travel times), and the program receives the values only when it runs, as a dictionary \texttt{VALUES} (Appendix~\ref{app:prompts}, Listing~\ref{lst:deter}). The model must therefore write out the problem's structure in placeholder names. Placeholder numbers are shuffled, so their order reveals nothing about the times. The model can still choose among placeholders (``meet Person2 first, from T3''), which the A judge counts as SLB. We apply placeholders only to Calendar and Meeting Planning, whose values separate cleanly from the structure. We run both methods for Qwen and GPT-OSS on the same 100 problems of every dataset and FL except PDDL, with $K=5$, and add $K \in \{2, 10\}$ for GPT-OSS on Calendar and Meeting Planning in Python. A batch's trace serves $K$ problems, so each of them is credited with the trace's ratio and $1/K$ of its tokens. The A judge overcounts SLB in placeholder traces, which can only understate the placeholders' effect (Appendix~\ref{app:ontology-judges}).

\textbf{Placeholders all but remove solver-like behavior without costing accuracy} (Figure~\ref{fig:deter}). Solver-like tokens per problem fall by 97\% $[96, 98]$, from 2{,}820 to 78; the ratio falls from 0.25 to 0.005, and the share of traces with SLB from 94\% to 15\%, for both models in both FLs. On the same problems, the cautionary prompts cut solver-like tokens by only 32\% and 18\%. Accuracy rises by 2.6 points $[-0.5, +5.9]$. The largest gain, GPT-OSS on Meeting Planning in Python (0.66 to 0.82), comes partly from programs that inherited the reasoning answer: 13 of the 28 problems that placeholders fix had such a program under the baseline. Since the model never sees a value, it cannot hardcode an answer, and the program, not the trace, computes the result. The cost is length for Qwen, whose reasoning per problem grows by 54\% (more than threefold on Calendar); in the longest of these traces we read, the added text concerns the program's robustness to values it cannot see, such as intervals ending at 24:00, not the solution. GPT-OSS's reasoning shrinks by 31\%.

\textbf{Batching shortens all reasoning, not just the solving, and costs accuracy.} Solver-like tokens per problem fall by 62\% $[55, 68]$, but reasoning tokens fall almost as much (61\%), so the ratio barely moves ($-0.02$ $[-0.045, +0.001]$; Qwen $+0.04$): the encoding is designed once for the batch, while each problem is still worked out in prose. Accuracy falls by 5.9 points $[-11.0, -0.5]$, most on ZebraLogic, by 15 to 26 points for each model in each FL. Larger batches cut more: for GPT-OSS on Calendar and Meeting Planning in Python, solver-like tokens per problem fall by 56\%, 78\% and 91\% at $K = 2$, 5 and 10, with accuracy 2, 6 and 3 points above the baseline.

\section{Conclusion}

Formalizers are meant to leave the search to a solver, yet their reasoning traces routinely do it themselves. Across three models and four datasets, this solver-like behavior is common and redundant, is not associated with accuracy, and depends most on the formal language. The few programs that inherit the reasoning answer also inherit its errors, and output-based evaluation cannot tell them from programs that search, so formalizers should be judged by how their programs are reached, not only by what they output. For practitioners, the most reliable remedy is to keep the instance's values out of the prompt: placeholders bound at run time remove nearly all solver-like behavior without loss of accuracy and rule out hardcoded answers. Otherwise, cautionary prompts remove only a quarter of it, and batching costs accuracy.

\section*{Limitations}

Our study covers three reasoning models and four constraint satisfaction datasets, each in two formal languages. Other task types, such as theorem proving or optimization, and other languages may show different amounts of solver-like behavior. DeepSeek is run in only one formal language per dataset because the model behind its API changed between our runs, so H4 rests on Qwen and GPT-OSS, and its runs cannot be reproduced exactly. The two prompts of a dataset differ in more than the formal language (\S\ref{sec:formulation}), notably the reference problem file in the PDDL prompt, which may contribute to the differences H4 attributes to it. Each problem is generated once per setting, so our intervals do not capture sampling variation across generations.

Our ontology labels come from LLM judges and scripts. The judges agree well with reference labels (Table~\ref{tab:reliability}), but the reference samples are small (110 paragraphs for A1, 40 traces for A2, 60 programs for B) and come from one annotator, one B check compares against labels written by a second model and the other showed the human the judge's evidence, A1 was checked on paragraphs and not on whole traces, and A2 was checked with the help of a script and a second model because full traces are too long to annotate by hand. B and C cover non-random subsets: B only programs that run (1{,}791 of 2{,}000, which drops most of Qwen's crashing Z3 programs), and C only the 1{,}547 complete reasoning answers, of which C1 can compare 1{,}314 with the program's output. C also covers only reasoning answers that the task's parser reads, and for Meeting Planning an extracted reasoning answer sometimes combines two tentative itineraries, so its error rate is an upper bound. The solver-like ratio counts whole paragraphs, so a paragraph that mixes encoding and solving counts entirely toward it.

Finally, H3 is observational: fixed effects for the problem remove differences in difficulty, but not every confounder. The cautionary prompts in \S\ref{sec:deter} reduce solver-like behavior without changing accuracy, which supports H3, but they remove only part of it. Placeholders remove almost all of it, but apply only where a problem's values separate from its structure; we test them, and batching, on two models and not in PDDL, and placeholders on two datasets.

\section*{Ethical Considerations}

\textbf{Risks.} We study how formalizers reach their programs and see no direct path to misuse. The risk our results point to is in deployment: a program that hardcodes or inherits the reasoning answer looks as verifiable as one that searches, so systems that trust formalizer output without validating it may act on answers no solver checked (\S\ref{sec:results}). Solver-like reasoning is also redundant computation, which placeholders remove (\S\ref{sec:deter}).

\textbf{Data.} The datasets are synthetic English constraint satisfaction problems. Calendar Scheduling and Meeting Planning refer to fictional people by first name only; the problems contain no information about real people, and we found no offensive content in them or in the traces we read. Datasets, models and licenses are listed in Appendix~\ref{app:exp}.

\textbf{Human annotation.} All human reference labels (Table~\ref{tab:reliability}, Appendix~\ref{app:ontology-judges}) were written by one of the authors, an NLP researcher, using the definitions in the judge prompts. No crowdworkers or other participants were involved, and no personal data was collected.

\textbf{Use of AI assistants.} We used Claude Code as a coding and writing assistant: to write and debug the experiment and analysis code, to run experiments, and to draft and edit text. The authors designed the study, directed and checked this work, and take responsibility for the paper; all reported numbers come from scripts in our repository.

\bibliography{anthology, custom}

\appendix

\section{Experimental Details}
\label{app:exp}

Each model receives one zero-shot prompt per problem, built from a prefix for the formal language and an output format for the task (Appendix~\ref{app:prompts}). We extract the program from the model's final answer and run it: Python and Z3 programs with the Python interpreter, and PDDL domains and problems with the Fast Downward planner. The output is scored as in Appendix~\ref{app:tasks}, and each generation is labeled by the ontology judges (Appendix~\ref{app:ontology-judges}). Each dataset contributes the same 100 randomly sampled problems in both formal languages. Table~\ref{tab:runs} lists every run: 6{,}000 generations, of which the 2{,}000 under the baseline prompt are used in \S\ref{sec:results}. It also counts the programs flagged as not written in the requested formal language: those that import a library outside the standard library in Python, or Z3 programs that never import Z3 or never call the solver. PDDL is not checked, since a planner rejects anything else.

\textbf{Artifacts and licenses.} Calendar Scheduling and Meeting Planning come from NaturalPlan \cite{zheng2024naturalplanbenchmarkingllms} (code Apache 2.0, data CC BY 4.0), ZebraLogic from \citet{lin2025zebralogic} (released for evaluation, without a stated license), and BlocksWorld from the IPC domain \cite{ipc}. All are evaluation benchmarks, and we use them only for evaluation. GPT-OSS-120B, Qwen3.8-27B and Gemma~4~31B-it are released under Apache 2.0 and run locally; DeepSeek-V4-Flash is used through its API under DeepSeek's terms of service. We release our code, prompts, model outputs and labels under the MIT license for research on formalizers.

\textbf{Software and compute.} Local models and judges run on vLLM 0.32.0 (PyTorch 2.13), one NVIDIA H100 NVL (94\,GB) per model instance; GPT-OSS-120B uses its released MXFP4 weights and the Gemma judges FP8. Python and Z3 programs run under Python 3.12 with z3-solver 4.14--4.15 and a 30-second limit. PDDL is solved by Fast Downward (commit \texttt{487e33c}, alias \texttt{lama-first}), and plans are validated by VAL (commit \texttt{3c7a1f3}). Tokens are counted with tiktoken's \texttt{cl100k\_base}. Generation and judging, including judge development and \S\ref{sec:deter}, took roughly 150 H100 GPU-hours by our logs, plus the DeepSeek API calls. Decoding settings (Appendix~\ref{app:models}) are fixed for all runs; we did no hyperparameter search.

\begin{table}[t]
\centering
\small
\setlength{\tabcolsep}{4pt}
\input{tables/runs.tex}
\caption{Runs: each model, task and formal language has 300 generations (100 problems under each of three prompts; 6{,}000 in total). Not in FL: programs not written in the requested FL under the baseline (0) and Cautionary A and B prompts. DeepSeek is run in one FL per dataset.}
\label{tab:runs}
\end{table}

\section{Tasks and Datasets}
\label{app:tasks}

Each dataset consists of constraint satisfaction problems (CSPs) \cite{russell2016artificial} stated in natural language, from which we sample 100 problems with their ground-truth solutions. Every dataset is formalized in Python, restricted to the standard library, and in a declarative formal language: Z3 for Calendar Scheduling, Meeting Planning and ZebraLogic, and PDDL \cite{mcdermott1998pddl} for BlocksWorld.

\textbf{Calendar Scheduling} \cite{zheng2024naturalplanbenchmarkingllms} involves scheduling meetings among multiple people given their existing constraints and availability windows. Models must find time slots that satisfy duration requirements and avoid conflicts with busy intervals for all participants.

\textbf{Meeting Planning} \cite{zheng2024naturalplanbenchmarkingllms} focuses on coordinating multi-location meetings with logistical constraints including people, locations, available time windows, and travel time between locations. Solutions require planning feasible itineraries that respect temporal and spatial constraints.

\textbf{ZebraLogic} \cite{lin2025zebralogic} presents logic grid puzzles where models must determine which attributes (e.g., colors, pets, drinks) belong to which entities (e.g., houses, people) based on a set of logical clues. In Z3, the program states the clues as constraints over the attributes; in Python, it must search the assignments itself, typically by enumerating permutations.

\textbf{BlocksWorld} \cite{ipc} is classical planning: rearranging blocks into a goal configuration under additional constraints. In PDDL, the model writes a domain and a problem, given a reference problem that omits the additional constraint (in 53 of the 100 instances it therefore differs from the ground-truth problem), and the Fast Downward planner finds the plan; in Python, the program prints the plan itself. Either plan is checked by the VAL validator against the ground-truth domain and problem.

\textbf{Accuracy.} For each problem instance, we construct a constraint JSON containing relevant restrictions and verify that the program's output violates none of them. Since multiple valid solutions may exist for some of the tasks (e.g. Calendar Scheduling), a generation is marked correct if the program executes successfully and its output satisfies all constraints, and incorrect otherwise. For Meeting Planning, the itinerary is simulated step by step against the start location and time, the travel times, and each friend's location, availability window, and minimum meeting duration; since the goal is to meet as many friends as possible, it must also meet at least as many friends as the ground-truth solution. For ZebraLogic, the printed grid must match the ground-truth solution. For BlocksWorld, the plan, found by the planner for a PDDL formalization or printed by a Python program, must be valid under the ground-truth domain and problem, and for unsolvable instances the formalization must be proven unsolvable or the program must report that no plan exists. Ten of the 100 BlocksWorld instances are unsolvable; for one of them the planner cannot prove unsolvability within its time limit even from the ground-truth PDDL, so PDDL accuracy on BlocksWorld is at most 99\%.

\section{Model Details}
\label{app:models}

We evaluate on three models with varying reasoning capabilities. All models support extended reasoning modes that produce explicit reasoning traces before generating code, making them suitable for analyzing solver-like behavior.

\textbf{DeepSeek-V4-Flash} \cite{deepseekai2025deepseekv32pushingfrontieropen} is a large reasoning model accessed via the DeepSeek API with thinking enabled, reasoning effort \emph{medium}, and at most 131{,}072 output tokens.

\textbf{GPT-OSS-120B} \cite{openai2025gptoss120bgptoss20bmodel} is a 120-billion-parameter model run locally with vLLM at reasoning effort \emph{medium} (temperature 0.7, top-$p$ 0.9), generating up to the 131{,}072-token context window.

\textbf{Qwen3.8-27B} \cite{qwen3technicalreport} is a 27-billion-parameter model run locally with vLLM with thinking enabled (temperature 0.7, top-$p$ 0.9), generating up to the 131{,}072-token context window.

All models are evaluated as formalizers, prompted to generate executable Python code (or Z3/PDDL, where appropriate) that returns the solution upon execution.

\textbf{DeepSeek in one FL.} DeepSeek is run in one FL per dataset: Python for Calendar Scheduling and Meeting Planning, Z3 for ZebraLogic, and PDDL for BlocksWorld. The model served under its API name changed before our runs in the other FL: re-running 20 earlier prompts with the same settings gives traces 4--5 times shorter, on every problem. Since its runs in the other FL would compare two different models, we leave them out.

\section{Prompt Configurations}
\label{app:prompts}

We evaluate each model under three prompting conditions to assess whether explicit warnings can deter solver-like behavior. All prompts consist of the same two parts: (i) a prefix for the formal language (Python, Z3 or PDDL) that instructs the model to formalize the problem, and (ii) a task-specific output format, which for Z3 also gives the import and the coding steps.

\textbf{Baseline} is the default prompt. The Python and Z3 prefixes instruct the model to write a program that ``solves it directly rather than solving it yourself in prose.'' The baseline discourages solving in natural language only implicitly. 
The PDDL prefix has no such clause; it supplies the ground-truth problem PDDL as a reference for the initial and goal states, and asks for a domain and a problem that encode the problem's constraints. The baseline measures how much solver-like behavior persists under an ordinary formalization instruction.

\textbf{Cautionary A} frames solver-like behavior by its outcome: the model works out the answer in its reasoning trace, and the code then just outputs that precomputed result. It contrasts a solver-like and a sound response to a multi-step wage problem with a conditional bonus; the sound response lists the given values and enumerates the computation steps without performing any arithmetic. It closes with an explicit instruction not to compute intermediate values or final answers during reasoning.

\textbf{Cautionary B} defines solver-like behavior by the act itself: substituting the given values into arithmetic, or deriving an intermediate or final result in the trace, before any code is written. It illustrates this with a single-step GSM8K problem, and its sound example refers to the given values only abstractly. It closes by instructing the model to reason only about the variables and constraints and to outline the steps a program would take.

Listings~\ref{lst:baseline}--\ref{lst:cautb} show the three prompts in Python for Calendar Scheduling; Listing~\ref{lst:z3pddl} shows the Z3 prefix with its Calendar Scheduling output format, and the PDDL prefix. The cautionary prompts add the same warning to every prefix.

Both cautionary prompts use math word problems rather than benchmark instances, so they leak no task-specific solutions but require the model to transfer the contrast to scheduling, logic, and planning. This design allows us to test whether explicit prompt variations reduce solver-like behavior and whether such reductions change formalization performance (\S\ref{sec:deter}).

\textbf{Batching and placeholders} (\S\ref{sec:deter}) change the problem part of the baseline prompt instead of adding a warning. A batched prompt uses the baseline prefix in the plural (``You are given $K$ constraint satisfaction problems (CSPs) of the same kind \dots{} enclose each problem's complete solution in its own code block''), lists the problems under headers ``\#\#\# Problem $i$'', gives the output format once, and ends with the instruction at the top of Listing~\ref{lst:deter}. Each problem is assigned the program headed by its number. A placeholder prompt is the baseline prompt with the problem rewritten in placeholders and preceded by a note on them (Listing~\ref{lst:deter}). Before execution, the program is prefixed with the definition of \texttt{VALUES} (and, for Meeting Planning, \texttt{TRAVEL}), and placeholder names in its output are mapped back to their values. The A judge sees the placeholder problem, with a note that choosing among placeholder values counts as SLB. Batch groups are fixed by a seeded shuffle of the 100 problems, so every model and FL sees the same groups.

\section{Ontology Judges and Their Reliability}
\label{app:ontology-judges}

All LLM judges use Gemma~4~31B-it (FP8, temperature 0, thinking mode), which is not one of the evaluated models; their prompts were fixed before the labeling runs. Reference labels are sampled evenly across each judge's predicted labels, so rare labels such as \emph{hardcoded} are tested as often as common ones. Because the judges were developed on traces in the FLs DeepSeek is run in, A1 and B are also checked by a human on held-out traces in each dataset's other FL. Each judge's input, output and reliability check are as follows; Listings~\ref{lst:judge-a}--\ref{lst:judge-b} give their prompts.

\textbf{Segments and paragraphs.} The A1 judge labels paragraphs, a unit a human can also label, but a whole trace is too long for one request (a DeepSeek trace averages 26{,}572 tokens). We therefore cut each trace into segments of at most 2{,}000 tokens (counted with the \texttt{cl100k\_base} tokenizer). Cuts fall between paragraphs, i.e.\ at blank lines; only a paragraph longer than a whole segment is cut between sentences. Each segment is then divided into numbered units for labeling: its paragraphs, where a paragraph over 700 characters is cut at line breaks (and a very long line at sentence ends) into pieces of at most about 700 characters. One request holds the problem and one segment, and states which segment of how many it is; the judge returns one label per unit and sees no other segment, so the context of a unit is its own segment. We keep segments short because long ones made the judge loop: in a pilot with an earlier prompt, 6{,}000-token segments looped until the output budget ran out in 83\% of responses, against 7\% at 2{,}000 tokens. The 2{,}000 generations of \S\ref{sec:results} give 14{,}027 segments (median 3 per trace, at most 58; 31\% of traces fit in one) and 372{,}234 units (median 24 per segment, median length 153 characters). The 2{,}046 units (0.5\%) that the judge's response leaves unlabeled count as \emph{none}. The solver-like ratio is the share of a trace's characters in units labeled \emph{SLB}, over all its segments. The other judges read one request per trace. The A2 judge reads the trace condensed to its \emph{SLB} units, in order, with ``[\ldots]'' marking each run of units left out. The B judge reads the whole trace; the two traces longer than 400{,}000 characters keep their first 50{,}000 and last 350{,}000.

\textbf{A1 (existence).} The judge labels every paragraph of a segment, given the problem. Its prompt distinguishes paragraphs that establish values (V), argue a property of the whole solution set such as uniqueness (P), both (VP), or neither (N); V and VP are \emph{SLB}, and P and N are \emph{none}. The prompt defines the decision variables of each task and lists what is not solver-like (restating, encoding design, checking the encoding on a hypothetical example, assuming solvability, and one-off checks of what a constraint means). Three rounds of human labels (60 paragraphs, the third balanced by the judge's predictions) plus 10 paragraphs selected to probe arguments about the solution set give 70 references. The judge agrees on 66/70 (49/52 on paragraphs not paraphrased in its prompt), with precision and recall of \emph{SLB} of 20/22 each. In the other FL of each dataset (the one DeepSeek is not run in), a human labeled 40 paragraphs blind (10 per dataset, half of them predicted \emph{SLB}, one per trace): the judge agrees on 37/40 (\emph{SLB} precision 18/20, recall 18/19; Table~\ref{tab:reliability} reports these blind labels), and on 39/40 after the annotator reviewed the disagreements, with precision 19/20 and recall 19/19. The remaining miss is a boundary case: a paragraph that mostly interprets a soft preference but ends by choosing between two candidate answers found earlier.

\textbf{A2 (extent).} The judge reads only the paragraphs labeled \emph{SLB}, condensed into one text, and the program's output, and states whether they contain a complete answer, which it writes in the task's output format. Because the traces are too long for annotators to read in full, the label \emph{complete} was verified by humans, and \emph{partial} labels were checked by searching the trace for the program's answer and by a second model reading the passages found. The judge agrees on 20/20 traces in the first round and 19/20 in the second (the miss follows from an A1 error).

\textbf{B (program).} The judge sees the problem, the trace (its beginning and end when long) and the program with its output, and returns a label with the evidence for it; a script sets B to n/a when the program fails or prints nothing. Only decisions the trace had to solve for count as inherited: a restriction read off the problem statement, or a preference encoded as a hard constraint, leaves the program independent. On 20 programs balanced by predicted label, the judge agrees with blind reference labels on 19. These reference labels were written by a second model, and the prompt's examples were drawn from them, so this agreement is optimistic. On 20 held-out programs that the judge labels \emph{hardcoded} (10) or \emph{partly fixed} (10), a human shown the judge's evidence agreed with all 20 labels. These programs are concentrated in two cells: every hardcoded one is GPT-OSS on Meeting Planning, where hardcoding is common, and 9 of the 10 partly fixed ones are BlocksWorld. In the other FLs, a human shown the judge's evidence checked 20 further programs, 9 of them in Z3: 10 that the judge labels B1--B3 (3 each of \emph{hardcoded} and \emph{steered}, 4 \emph{partly fixed}) and, as a check of recall, 10 that it labels \emph{independent} although the trace reached a complete answer, shown next to the program. The human agrees on 18/20: all 10 \emph{independent} labels, and 8 of the 10 others. In the two misses the judge took a program detail for inheritance: an action order labeled \emph{steered} cannot change the result, because the shortest plan is unique, and a \emph{partly fixed} name order is the problem's listing order, left in by a missing loop.

\textbf{Placeholder traces.} The A judge was developed on traces that name concrete values. In placeholder traces (\S\ref{sec:deter}), a human read 20 paragraphs it labels \emph{none} that mention placeholders alongside words such as \emph{first} or \emph{earliest}: all 20 interpret the problem or design the code, so none is a missed SLB paragraph. Of 10 random paragraphs it labels \emph{SLB}, the human judged 4 solver-like, such as one that orders the placeholder times and sweeps two calendars for a free slot; the other 6 reason about the program's behavior on unseen values or interpret a constraint. The judge thus overcounts SLB in placeholder traces, so the reduction placeholders achieve is, if anything, understated.

\textbf{C (output).} C1 and C2 are deterministic scripts given the reasoning answer extracted for A2: C1 parses the reasoning answer and the program's output with the task's parsers and compares them, and C2 runs the task's validator on the reasoning answer. A reasoning answer that the task's parser cannot read makes C1 n/a, and we report how often this happens.

\textbf{Checks over all traces.} Two checks over all traces agree with the reference comparisons. Of the 1{,}314 complete reasoning answers that can be compared with the program's output, 88\% match it exactly, which an A2 extraction that invented answers would not achieve. The A and B judges run independently, yet only 6 of the 163 programs labeled B1--B3 come from traces without \emph{SLB}, the combination the ontology rules out.

\section{Full Results}
\label{app:full-results}

Table~\ref{tab:ontology-cells} lists the ontology labels of every run, in both formal languages and under all three prompts. It is generated from the per-generation labels by \texttt{scripts/paper\_figures.py}.

\begin{table*}[p]
\centering
\setlength{\tabcolsep}{3pt}
\adjustbox{max width=\textwidth, max totalheight=0.84\textheight}{\input{tables/ontology_cells.tex}}
\caption{Ontology labels per cell (100 generations). FL: formal language (DeepSeek: one per dataset). Cond.: 0 = Baseline, A and B = Cautionary A and B. Acc.: program accuracy. SLB: \% of traces with SLB (A1). Cmpl., Part.: \% of traces with a complete or partial reasoning answer (A2). Ratio: mean solver-like ratio. SLB tok.: mean solver-like tokens per trace. B1--B4, n/a: \% of programs. C1 no: reasoning answers that differ from the program's output, of those compared. C2 no: wrong reasoning answers, of those checked.}
\label{tab:ontology-cells}
\end{table*}

\section{Labeled Examples}
\label{app:examples}

Figures~\ref{fig:example-a} and~\ref{fig:example-b} show real outputs with the labels of the final judges (\S\ref{sec:formulation}, Appendix~\ref{app:ontology-judges}). Text is quoted verbatim except for elisions marked [\ldots] and the GPT-OSS channel markers, which we remove.

\newtcolorbox{exvalues}[1]{colback=cDS!14, colframe=cDS!80!black, boxrule=0.4pt, arc=2pt, left=3pt, right=3pt, top=1.5pt, bottom=1.5pt, before skip=2pt, after skip=2pt, fontupper=\footnotesize, before upper={\textsf{\textbf{\color{cDS!70!black}#1 SLB}}\enspace}}
\newtcolorbox{exother}[1]{colback=black!3, colframe=black!40, boxrule=0.4pt, arc=2pt, left=3pt, right=3pt, top=1.5pt, bottom=1.5pt, before skip=2pt, after skip=2pt, fontupper=\footnotesize, before upper={\textsf{\textbf{\color{black!60}#1 none}}\enspace}}

\begin{figure*}[p]
\centering
\begin{minipage}{0.97\linewidth}
{\footnotesize\textbf{Problem} (\texttt{calendar\_scheduling\_example\_32}, GPT-OSS-120B, Baseline): schedule a 30-minute meeting for Emily, Melissa and Frank between 9:00 and 17:00 on Monday. Emily is busy 10:00--10:30, 11:30--12:30, 14:00--15:00, 16:00--16:30; Melissa 9:30--10:00, 14:30--15:00; Frank 10:00--10:30, 11:00--11:30, 12:30--13:00, 13:30--14:30, 15:00--16:00, 16:30--17:00. Frank does not want to meet on Monday after 9:30.}
\medskip

\begin{exother}{1}We need to parse the problem. We have to formalize variables and constraints, then write python code that solves it. The task description: schedule a meeting for Emily, Melissa and Frank for half an hour between work hours 9:00 to 17:00 on Monday. Existing schedules given. Additional constraint: Frank do not want to meet on Monday after 9:30. Means meeting must start no later than 9:30? Actually ``after 9:30'' means after 9:30 (i.e., start after 9:30?), but Frank does not want to meet after 9:30. So meeting must be at or before 9:30 start, and duration 30 minutes, so meeting must be within 9:00-9:30? Let's see.\end{exother}
\begin{exother}{2}Work hours 9:00-17:00. Frank busy intervals: 10:00-10:30, 11:00-11:30, 12:30-13:00, 13:30-14:30, 15:00-16:00, 16:30-17:00.\end{exother}
\begin{exvalues}{3}Frank does not want to meet after 9:30, meaning meeting must finish by 9:30? [\ldots] So meeting must be within 9:00-9:30 (start at 9:00, end at 9:30). Let's check other participants: Emily has meetings 10:00-10:30 etc, none before 10:00, so she is free 9:00-10:00. Melissa is busy 9:30-10:00, so she is free 9:00-9:30. So the only common free slot before constraints is 9:00-9:30. That fits Frank's preference.\end{exvalues}
\begin{exvalues}{4}Thus answer: \texttt{['Monday', '09:00', '09:30']}\end{exvalues}
\begin{exother}{5}We need to produce code that reads nothing (hardcode data) and outputs that list. Use constraints solving approach maybe brute force intervals.\end{exother}
\begin{exother}{6}We'll parse intervals as minutes from midnight. Define work start end. Build busy intervals for each participant. Add constraint that meeting duration 30 minutes, start >= work\_start, end <= work\_end, and must not intersect any busy interval of any participant. Also constraint Frank preference: meeting end <= 9:30? [\ldots] The earlier reasoning gave unique solution 9:00-9:30. We'll enforce meeting end <= 9:30 to be safe.\end{exother}
\begin{exother}{7}Implement algorithm: iterate start minute from work\_start to work\_end - duration inclusive. For each start, end=start+duration. Check constraints and break when found. Print list with formatting.\end{exother}
\begin{exother}{8}Let's write code accordingly.\end{exother}
\medskip

{\footnotesize\textbf{Program} (excerpt; it prints \texttt{['Monday', '09:00', '09:30']}, scored correct):}
\begin{lstlisting}[language=Python, basicstyle=\ttfamily\scriptsize, columns=fullflexible, aboveskip=2pt, belowskip=2pt]
frank_latest_end = to_min("09:30")
def feasible(start):
    end = start + duration
    if start < work_start or end > work_end: return False
    if end > frank_latest_end: return False      # Frank's preference
    for person, intervals in busy_min.items():
        for b_start, b_end in intervals:
            if overlaps(start, end, b_start, b_end): return False
    return True
for minute in range(work_start, work_end - duration + 1):
    if feasible(minute):
        solution = [day, to_str(minute), to_str(minute + duration)]; break
\end{lstlisting}

{\footnotesize\textbf{Labels.} \textbf{A1} SLB (paragraphs 3--4). \textbf{A2} complete: Monday 9:00--9:30. \textbf{Solver-like ratio} 0.28. \textbf{B4} independent; judge's evidence: ``The program encodes all provided busy schedules and the preference constraint (`end <= 9:30') and searches for a feasible slot using a loop.'' \textbf{C1} yes, \textbf{C2} yes.}
\end{minipage}
\caption{A trace with every paragraph and its A1 label. Paragraph~1 only interprets Frank's preference, which is not solver-like; paragraph~3 checks the schedules and rules out every slot but one. Although the trace has already found the unique answer, the program searches for it independently (B4), so the SLB is wasted.}
\label{fig:example-a}
\end{figure*}

\begin{figure*}[p]
\centering
\begin{minipage}{0.97\linewidth}
\begin{tcolorbox}[colback=white, colframe=cOSS!70!black, boxrule=0.6pt, arc=2pt, left=4pt, right=4pt, top=2pt, bottom=2pt, title={\textsf{B1 Hardcoded} \hfill {\normalfont\texttt{meeting\_planning\_example\_66}, GPT-OSS-120B, Baseline}}, coltitle=black, colbacktitle=cOSS!15, fonttitle=\footnotesize\bfseries]
{\footnotesize\textit{Trace:} ``Calculate times: 9:00 AM start. Wait until 10:58 AM departure. Travel 17 min, arrive at 11:15 AM (since 10:58+17=11:15). Meet from 11:15 to 1:15 (120 min). That satisfies.''}
\begin{lstlisting}[language=Python, basicstyle=\ttfamily\scriptsize, columns=fullflexible, aboveskip=3pt, belowskip=1pt]
itinerary = [
    "You start at Nob Hill at 9:00AM.",
    "You wait until 10:58AM.",
    "You travel to Presidio in 17 minutes and arrive at 11:15AM.",
    "You meet Robert from 11:15AM to 1:15PM."
]
print(itinerary)
\end{lstlisting}
{\footnotesize\textit{Judge:} ``The program defines a list `itinerary' containing the final answer strings derived in the prose and prints it without performing any computation.''}
\end{tcolorbox}

\begin{tcolorbox}[colback=white, colframe=cOSS!70!black, boxrule=0.6pt, arc=2pt, left=4pt, right=4pt, top=2pt, bottom=2pt, title={\textsf{B2 Partly fixed} \hfill {\normalfont\texttt{meeting\_planning\_example\_73}, GPT-OSS-120B, Baseline}}, coltitle=black, colbacktitle=cOSS!15, fonttitle=\footnotesize\bfseries]
{\footnotesize\textit{Trace:} ``We can travel from Russian Hill to Pacific Heights (7 min) arriving at 9:07AM. Then we can meet from 9:07 to 10:07 (60 min) which is within Barbara's presence (till 10PM). That satisfies constraints. [\ldots]''}
\begin{lstlisting}[language=Python, basicstyle=\ttfamily\scriptsize, columns=fullflexible, aboveskip=3pt, belowskip=1pt]
start = datetime.datetime(2023, 1, 1, 9, 0)          # 9:00AM
travel_minutes = 7
meet_minutes = 60
arrive = start + datetime.timedelta(minutes=travel_minutes)
meet_end = arrive + datetime.timedelta(minutes=meet_minutes)
itinerary = [
    f"You start at Russian Hill at {fmt(start)}.",
    f"You travel to Pacific Heights in {travel_minutes} minutes and arrive at {fmt(arrive)}.",
    f"You meet Barbara from {fmt(arrive)} to {fmt(meet_end)}."
]
print(itinerary)
\end{lstlisting}
{\footnotesize\textit{Judge:} ``The program implements a specific itinerary decided in the prose, using a fixed meeting duration of 60 minutes and a pre-determined sequence of events.''}
\end{tcolorbox}

\begin{tcolorbox}[colback=white, colframe=cOSS!70!black, boxrule=0.6pt, arc=2pt, left=4pt, right=4pt, top=2pt, bottom=2pt, title={\textsf{B3 Steered} \hfill {\normalfont\texttt{meeting\_planning\_example\_91}, Qwen3.8-27B, Baseline}}, coltitle=black, colbacktitle=cOSS!15, fonttitle=\footnotesize\bfseries]
{\footnotesize\textit{Trace:} ``[\ldots] likely expected: \texttt{['You start at Russian Hill at 9:00AM.', 'You travel to Richmond District in 14 minutes and arrive at 9:14AM.', 'You wait until 7:00PM.', 'You meet Daniel for 75 minutes from 7:00PM to 8:15PM.']}'' \ldots\ ``If we choose earliest departure (start at 9:00) and earliest meeting (7:00), output expected. That's a reasonable tie-breaker [\ldots]''}
\begin{lstlisting}[language=Python, basicstyle=\ttfamily\scriptsize, columns=fullflexible, aboveskip=3pt, belowskip=1pt]
for depart in range(start_time, 24 * 60):
    arrive = depart + travel_time
    earliest_meet = max(arrive, friend_start)
    latest_meet = friend_end - required_minutes
    if latest_meet < earliest_meet:
        continue
    for meet_start in range(earliest_meet, latest_meet + 1):
        key = (0, depart, meet_start, arrive, meet_start - arrive)
        if best is None or key < best[0]:
            best = (key, depart, arrive, meet_start)
\end{lstlisting}
{\footnotesize\textit{Judge:} ``The program uses a tie-breaking key \texttt{(0, depart, meet\_start, arrive, meet\_start - arrive)} to ensure the search returns the earliest possible departure and meeting time, a choice the reasoning explicitly states was made to reproduce the specific itinerary derived in the prose.''}
\end{tcolorbox}
\end{minipage}
\caption{Programs labeled B1--B3, each with an excerpt of the solver-like trace it follows and the B judge's evidence. All three are scored correct. The hardcoded program only prints the itinerary from the trace; the partly fixed one computes the times but along the single itinerary chosen in the trace; the steered one searches all departures and meeting times, but with a tie-break chosen so that the search returns the itinerary the trace expected.}
\label{fig:example-b}
\end{figure*}

\clearpage

\begin{listing*}[t]
\caption{Baseline Prompt: Universal CSP formalization prompt with no cautionary warnings, shown with the Calendar Scheduling output specification.}
\label{lst:baseline}
\begin{tcolorbox}[
    colback=white,
    colframe=black,
    boxrule=0.5pt,
    width=\linewidth,
    arc=0pt,
    outer arc=0pt,
    left=6pt, right=6pt, top=6pt, bottom=6pt
]
\begin{lstlisting}[style=prompt, basicstyle=\ttfamily\small]
You are given a constraint satisfaction problem (CSP) expressed in natural language: a set of entities, constraints between them, and a goal state that must be reached without violating any constraint. Formalize the variables and constraints from the text, then write a Python program that solves it directly rather than solving it yourself in prose. Use only the Python standard library, enclose the complete solution in a single ```python code block, and have the program print only the final answer in the exact format specified below.

CRITICAL: Your Python code block must be a complete, executable script with no prose, explanations, or natural language mixed into the code. Any text inside the ```python block must be valid Python syntax (code or comments only). All reasoning, explanations, or commentary must appear OUTSIDE the code block. The code must run without syntax errors when executed directly.

{PROBLEM}

This problem is a calendar-scheduling task: find a single meeting time that fits every participant's availability.

Print only a Python list of three strings — day, start time, end time.

Rules: the day must be a full weekday name (e.g. "Monday"), each time must be in zero-padded 24-hour HH:MM format with no AM/PM suffix and no seconds, and the list should follow the exact form [day, start time, end time] with no extra strings.

Here is an example:

['Monday', '13:30', '14:00']

SOLUTION CODE:
\end{lstlisting}
\end{tcolorbox}
\end{listing*}

\begin{listing*}[tp]
\small
\caption{Cautionary Prompt A: The Baseline prompt with an added outcome-focused warning against solver-like reasoning, illustrated with a multi-step problem containing a conditional. Shown with the Calendar Scheduling output specification.}
\label{lst:cauta}
\begin{tcolorbox}[
    colback=white,
    colframe=black,
    boxrule=0.2pt,
    width=0.99\linewidth,
    arc=0pt,
    outer arc=0pt,
    left=6pt, right=6pt, top=0pt, bottom=0pt
]
\begin{lstlisting}[style=prompt, basicstyle=\ttfamily\small]
You are given a constraint satisfaction problem (CSP) expressed in natural language: a set of entities, constraints between them, and a goal state that must be reached without violating any constraint. Formalize the variables and constraints from the text, then write a Python program that solves it directly rather than solving it yourself in prose. Use only the Python standard library, enclose the complete solution in a single ```python code block, and have the program print only the final answer in the exact format specified below.

CRITICAL: Your Python code block must be a complete, executable script with no prose, explanations, or natural language mixed into the code. Any text inside the ```python block must be valid Python syntax (code or comments only). All reasoning, explanations, or commentary must appear OUTSIDE the code block. The code must run without syntax errors when executed directly.

IMPORTANT: Avoid solver-like reasoning patterns. Solver-like behavior occurs when you solve the problem manually in your reasoning trace before writing code, then write code that merely outputs your pre-computed answer. This defeats the purpose of formalization.

Example problem: "Sarah earns $18 per hour tutoring students. Yesterday, she tutored for 45 minutes. Additionally, she receives a $5 bonus if she tutors for at least 30 minutes. How much did she earn in total?"

Solver-like reasoning (avoid this): Sarah earns $18 per hour. For 45 minutes, I need to convert this to a per-minute rate. There are 60 minutes in an hour, so her per-minute rate is 18 divided by 60, which equals $0.30 per minute. For 45 minutes of work, she earned 45 times 0.30, which is $13.50. Since she worked at least 30 minutes, she gets the $5 bonus. Adding the bonus: 13.50 plus 5 equals $18.50. Therefore, Sarah earned $18.50 total.

Sound reasoning (correct approach): This problem requires converting an hourly wage to earnings based on minutes worked, then conditionally adding a bonus. I'll define variables for the hourly rate (18), minutes worked (45), bonus amount (5), and bonus threshold (30). The formalization involves: (1) computing the per-minute rate by dividing the hourly rate by 60, (2) multiplying the per-minute rate by minutes worked to get base earnings, (3) checking if minutes worked meets the threshold, and (4) adding the bonus if the condition holds. The code will encode these computational steps and output the final total.

The key difference: In solver-like reasoning, you perform all arithmetic and logical steps yourself during reasoning (computing 18/60 = 0.30, then 45 * 0.30 = 13.50, then 13.50 + 5 = 18.50), arriving at the specific numerical answer before any code exists. The code then becomes a trivial output statement. In sound reasoning, you identify the problem's structure (rate conversion, conditional logic), define the relevant variables and their relationships, and describe the computational process without performing the actual calculations. Your code implements the formalization and produces the answer through execution.

Do not compute intermediate values or final answers during reasoning. Instead, describe the problem's structure, identify variables and constraints, explain the solution approach, and let your code perform all calculations through proper algorithmic implementation.

{PROBLEM}

This problem is a calendar-scheduling task: find a single meeting time that fits every participant's availability.

Print only a Python list of three strings — day, start time, end time.

Rules: the day must be a full weekday name (e.g. "Monday"), each time must be in zero-padded 24-hour HH:MM format with no AM/PM suffix and no seconds, and the list should follow the exact form [day, start time, end time] with no extra strings.

Here is an example: ['Monday', '13:30', '14:00']

SOLUTION CODE:
\end{lstlisting}
\end{tcolorbox}
\end{listing*}

\begin{listing*}[tp]
\caption{Cautionary Prompt B: The Baseline prompt with an added act-focused definition of solver-like behavior, illustrated with a single-step GSM8K problem. Shown with the Calendar Scheduling output specification.}
\label{lst:cautb}
\begin{tcolorbox}[
    colback=white,
    colframe=black,
    boxrule=0.5pt,
    width=\linewidth,
    arc=0pt,
    outer arc=0pt,
    left=6pt, right=6pt, top=6pt, bottom=6pt
]
\begin{lstlisting}[style=prompt, basicstyle=\ttfamily\small]
You are given a constraint satisfaction problem (CSP) expressed in natural language: a set of entities, constraints between them, and a goal state that must be reached without violating any constraint. Formalize the variables and constraints from the text, then write a Python program that solves it directly rather than solving it yourself in prose. Use only the Python standard library, enclose the complete solution in a single ```python code block, and have the program print only the final answer in the exact format specified below.

CRITICAL: Your Python code block must be a complete, executable script with no prose, explanations, or natural language mixed into the code. Any text inside the ```python block must be valid Python syntax (code or comments only). All reasoning, explanations, or commentary must appear OUTSIDE the code block. The code must run without syntax errors when executed directly.

We define solver-like behavior as when, instead of formalizing the problem's structure for code to solve, the reasoning performs the problem instance's actual computation in natural language - substituting the given values into arithmetic or deriving an intermediate or final result in prose - before any code is written.

For example, consider the following problem "Weng earns $12 an hour for babysitting. Yesterday, she just did 50 minutes of babysitting. How much did she earn?"

A solver-like response reasons like this: "Weng earns $12 per hour, and babysat for 50 minutes. Since there are 60 minutes in an hour, her per-minute rate is 12/60 = $0.2 per minute. For 50 minutes, she earned 50 * 0.2 = $10. So the answer is $10." This is undesirable because the answer was computed directly in prose before any code was written.

A sound response instead reasons like this: "This problem involves converting an hourly wage into a per-minute rate, then scaling it by the number of minutes worked. We can define variables for the hourly rate and minutes worked, convert the hourly rate to a per-minute rate by dividing by 60, and then multiply that per-minute rate by the number of minutes to get the total earnings. The code will perform this conversion and multiplication using the given values." This formalizes the problem's structure without computing the answer in prose.

Avoid the solver-like pattern. Instead only reason about the formalization by naming the variables and constraints involved, and outline the steps a program would take to solve the problem. Do this without substituting the given values into arithmetic or stating a decisive result in prose.

{PROBLEM}

This problem is a calendar-scheduling task: find a single meeting time that fits every participant's availability.

Print only a Python list of three strings — day, start time, end time.

Rules: the day must be a full weekday name (e.g. "Monday"), each time must be in zero-padded 24-hour HH:MM format with no AM/PM suffix and no seconds, and the list should follow the exact form [day, start time, end time] with no extra strings.

Here is an example:

['Monday', '13:30', '14:00']

SOLUTION CODE:
\end{lstlisting}
\end{tcolorbox}
\end{listing*}

\begin{listing*}[tp]
\caption{Batching and placeholders (\S\ref{sec:deter}). Top: the instruction that ends a batched prompt. Middle: the note that precedes a Calendar Scheduling problem with placeholders (Meeting Planning's note also introduces locations, durations and \texttt{TRAVEL}). Bottom: the task of a Calendar Scheduling problem with placeholders; the program receives \texttt{VALUES = \{"Person1": "Judy", "Person2": "Nicole", "Day1": "Monday", "T3": "9:00", ..., "D1": 30\}}.}
\label{lst:deter}
\begin{tcolorbox}[colback=white, colframe=black, boxrule=0.5pt, width=\linewidth, arc=0pt, outer arc=0pt, left=4pt, right=4pt, top=3pt, bottom=3pt]
\begin{lstlisting}[style=prompt]
Answer all {K} problems above. Give {K} ```python code blocks, one per problem, in the order of the problems. The first line of each block must be the comment `# Problem i`, where i is the problem's number. Each program must solve only its own problem and print only that problem's answer, in the format specified above. Do not combine problems into one program.
\end{lstlisting}
\medskip
\begin{lstlisting}[style=prompt]
The concrete values in this problem have been replaced by placeholders: Person1, Person2, ... for the participants, Day1, Day2, ... for weekdays, T1, T2, ... for clock times, and D1 for the meeting duration. Placeholder numbers carry no meaning; in particular, T1, T2, ... are not in chronological order. The values are the ones of an ordinary scheduling problem stated as written: every interval "Ta to Tb" ends after it starts, and all times lie within one day.

You are not told the values, and you do not need them: when your program runs, a dict named VALUES is already defined. VALUES contains exactly one entry per placeholder, mapping its name (a string such as "T1") to its value, and nothing else: no schedules, lists or constraints. Participants and weekdays are strings (e.g. "Monday"), clock times are 24-hour "H:MM" strings (e.g. "9:30" or "14:00"), and D1 is an integer number of minutes. Your program must therefore write out the problem's structure itself, in terms of placeholder names (for example busy = {"Person1": {"Day1": [("T1", "T4"), ("T7", "T2")]}} and the work hours and preferences), and look up each placeholder's value in VALUES (e.g. VALUES["T1"]). It must not define or overwrite VALUES, and must print the answer with the actual values (the real weekday and times), not the placeholder names.
\end{lstlisting}
\medskip
\begin{lstlisting}[style=prompt]
TASK: You need to schedule a meeting for Person1 and Person2 for D1 between the work hours of T3 to T5 on Day1. 

Here are the existing schedules for everyone during the day: 
Person1 is free the entire day.
Person2 has meetings on Day1 during T3 to T2, T4 to T6; 

Person2 would rather not meet on Day1 before T1. Find a time that works for everyone's schedule and constraints. 
\end{lstlisting}
\end{tcolorbox}
\end{listing*}

\begin{listing*}[p]
\caption{Baseline prompts in the declarative formal languages. Top: the Z3 prefix, followed by the Z3 output format for Calendar Scheduling, which gives the import and coding steps. Bottom: the PDDL prefix for BlocksWorld, which supplies the ground-truth problem PDDL as a reference; its output format asks for a domain and a problem as a JSON object.}
\label{lst:z3pddl}
\begin{tcolorbox}[colback=white, colframe=black, boxrule=0.5pt, width=\linewidth, arc=0pt, outer arc=0pt, left=4pt, right=4pt, top=3pt, bottom=3pt]
\begin{lstlisting}[style=prompt]
You are given a constraint satisfaction problem (CSP) expressed in natural language: a set of entities, constraints between them, and a goal state that must be reached without violating any constraint. Formalize the variables and constraints from the text, then write a Z3-Python program that solves it directly rather than solving it yourself in prose. You may use the z3-solver library (imported as `from z3 import *`), enclose the complete solution in a single ```python code block, and have the program print only the final answer in the exact format specified below.

CRITICAL: Your Python code block must be a complete, executable script with no prose, explanations, or natural language mixed into the code. Any text inside the ```python block must be valid Python syntax (code or comments only). All reasoning, explanations, or commentary must appear OUTSIDE the code block. The code must run without syntax errors when executed directly.

{PROBLEM}
\end{lstlisting}
\medskip
\begin{lstlisting}[style=prompt]
This problem is a calendar-scheduling task: find a single meeting time that fits every participant's availability.

Write Z3-Python code that declares variables, encodes all constraints, solves with a Z3 solver, and prints the meeting time.

Required imports:
```python
from z3 import *
```

Your code should:
1. Define Z3 variables for the meeting (day and start time)
2. Add domain constraints (work hours, meeting duration)
3. Encode all participant constraints (busy times, preferences)
4. Call `solver.check()` and retrieve the model
5. Print the solution in the format below

Print only a Python list of three strings — day, start time, end time.

Rules: the day must be a full weekday name (e.g. "Monday"), each time must be in zero-padded 24-hour HH:MM format with no AM/PM suffix and no seconds, and the list should follow the exact form [day, start time, end time] with no extra strings.

Here is an example:

['Monday', '13:30', '14:00']

SOLUTION CODE:
\end{lstlisting}
\medskip
\begin{lstlisting}[style=prompt]
You are given a constraint satisfaction problem (CSP) expressed in natural language: a set of entities, constraints between them, and a goal state that must be reached without violating any constraint. Formalize the variables and constraints from the text, then write PDDL code (Planning Domain Definition Language) that defines the planning domain and problem. PDDL is NOT Python - it is a specialized planning language with Lisp-like syntax.

CRITICAL: Your code block must contain valid PDDL 1.0 syntax only (with comments allowed using semicolons `;`). All reasoning, explanations, or commentary must appear OUTSIDE the code block. The PDDL must be syntactically correct and compatible with standard PDDL planners (VAL validator).

IMPORTANT CONTEXT:
The groundtruth problem PDDL for this specific instance (with the correct initial and goal states) is provided below. You should use this as a reference for:
- Understanding the exact initial configuration
- Understanding the exact goal configuration
- The correct object names and structure

However, you MUST create your own domain PDDL that properly encodes the constraints mentioned in the problem description.

GROUNDTRUTH PROBLEM PDDL (for reference):
```
{GROUNDTRUTH_PROBLEM_PDDL}
```

DOMAIN DESCRIPTION:
{DOMAIN_DESCRIPTION}

PROBLEM-SPECIFIC DESCRIPTION:
{PROBLEM_DESCRIPTION}

FULL PROBLEM WITH CONSTRAINTS:
{PROBLEM}
\end{lstlisting}
\end{tcolorbox}
\end{listing*}

\begin{listing*}[p]
\caption{Prompt of the A1 judge, applied to one segment of a trace at a time. V and VP are read as \emph{SLB}, P and N as \emph{none}. ``Prose'' in the judge prompts refers to the reasoning trace.}
\label{lst:judge-a}
\begin{tcolorbox}[colback=white, colframe=black, boxrule=0.5pt, width=\linewidth, arc=0pt, outer arc=0pt, left=4pt, right=4pt, top=3pt, bottom=3pt]
\begin{lstlisting}[style=prompt]
You are annotating the reasoning of a language model that was asked to write a program (Python, Python with Z3, or PDDL) for a constraint problem. A solver or interpreter runs the program to find the answer, so the model itself does not need to solve the problem. You will label each numbered paragraph of one excerpt of its reasoning.

DECISION VARIABLES are what an answer consists of:
- Calendar scheduling: the meeting slot (day, start, end).
- Meeting planning: which friends to meet, in what order, at what times.
- Zebra puzzles: which value of each attribute goes in which house.
- BlocksWorld: the plan (the sequence of actions).

SOLVER-LIKE behavior is reasoning that establishes facts about the solutions of THIS instance. Give each paragraph one TYPE: V, P, VP (both) or N.

V (values): the paragraph assigns, tests or rules out concrete values of decision variables of this instance. Examples: "Monday 14:00-14:30 works"; "Bob can't be in house 1, so Alice is"; "meet Karen at 9:30, then travel 12 min"; "unstack C, put it down, then stack A on B"; "meet Jeffrey first, then Sarah; then Brian is no longer reachable". Ruling out or narrowing candidates counts: "Natalie is busy until 11:30, so no slot before 11:30 works"; ruling out a friend whose window is too short; propagating that two attributes share a house; computing free intervals in order to find a slot.

P (properties) is narrow. A paragraph has the properties type only if BOTH of these hold:
  1. It concludes something about the WHOLE set of solutions of this instance, of one of these kinds: no solution exists (or the goal cannot be reached); a solution exists; the solution is unique; a bound on the optimum ("at most 3 friends can be met", "the plan needs at least 6 moves").
  2. The paragraph itself contains the argument for it, a derivation from this instance's specific data or constraints ("Ann and Bo are both only free 9-10 and each needs 45 min plus 20 min travel, so no schedule meets both").
VP (both): the paragraph does both, e.g. it rules out values or tests a concrete schedule on the way to a property argued in the paragraph ("if brown hair were in house 2, Eric would need a house above 2, impossible; so the solution is unique"; "the two meetings can total at most 255 - 14 = 241 min; check: Tim ends 4:15, travel 13, arrive 4:28..."). Use P only when no concrete values are committed, tested or ruled out.

A paragraph that claims such a property but fails 1 or 2 has no properties type (it is V if it also handles values, else N), in particular:
  - a property conclusion whose argument is in an earlier paragraph ("Thus the goal is unreachable.", "So max 8."). This applies ONLY to claims about the solution set: a paragraph that states or sums up concrete values ("so Ann, Bo and Cy, ending at 20:30, 6 friends") is V, even if they were derived earlier;
  - a conclusion with no argument ("all 9 can be met", "the solution is unique, good", "we found it unique");
  - assuming solvability, e.g. because the problem says a solution exists, or hedging ("it may be impossible, but we only need to encode it");
  - statements about the program or its search rather than the instance's solutions ("the search space is small, brute force is fine", "Z3 might return a different model if there are several").

N (not solver-like): everything else, including restating the problem's data or constraints (even with small rewrites such as converting times to minutes), designing the encoding or the code, choosing data structures, output format, parsing, writing or quoting code, and interpreting what a constraint means. Also N:
  - checking the program or encoding against the problem, e.g. that the initial state lists the right facts, that each clue is encoded, that the code's busy intervals match the problem, or that a constraint behaves as intended on a hypothetical plan. This is N even if the paragraph works out a value while checking ("so Arnold must be first; the code doesn't say that explicitly") or measures the instance for the encoding ("the tallest initial stack has 7 blocks, so heights up to 7 must be tracked");
  - interpreting a constraint, including a one-off check of one example value that was not obtained by solving ("does 11:30 count as morning? It should"). Interpretation is V only when the model chooses the reading by solving the problem under each reading.

Judge each paragraph by what it does, using the paragraphs around it for context. Short connective paragraphs ("Let's check.", "Now the code.") are N unless they themselves state a value.

OUTPUT FORMAT. One line per paragraph, in order, for every paragraph number, and nothing else:
<number>|<TYPE>
Example for a five-paragraph excerpt:
1|N
2|V
3|VP
4|P
5|N

PROBLEM GIVEN TO THE MODEL:
{problem}

REASONING EXCERPT ({n} paragraphs; excerpt {segment} of {n_segments} of the full reasoning):
{paragraphs}
\end{lstlisting}
\end{tcolorbox}
\end{listing*}

\begin{listing*}[p]
\caption{Prompt of the A2 judge, which extracts the reasoning answer from the paragraphs the A1 judge labeled \emph{SLB} and decides whether it is complete.}
\label{lst:judge-a2}
\begin{tcolorbox}[colback=white, colframe=black, boxrule=0.5pt, width=\linewidth, arc=0pt, outer arc=0pt, left=4pt, right=4pt, top=3pt, bottom=3pt]
\begin{lstlisting}[style=prompt]
A language model was asked to write a program for a constraint problem, and reasoned in prose before writing it. Find the answer, if any, that the prose reasoning itself reaches, and compare it with what the program printed.

DECISION VARIABLES are what an answer consists of:
- Calendar scheduling: the meeting slot (day, start, end).
- Meeting planning: which friends to meet, in what order, at what times.
- Zebra puzzles: which value of each attribute goes in which house.
- BlocksWorld: the plan (the sequence of actions).

EXTENT of the prose answer. Use the reasoning's FINAL position if it changes its mind. Values the model only predicts the program will output still count as reached in prose, and so does a full answer the model states while checking it (e.g. "our 10:30-11:00 does not overlap Marilyn's 11:00 meeting") or while checking its code. Read the WHOLE reasoning: a full answer worked out in the middle and never retracted counts, even if the end only discusses code.
- complete: every decision variable gets a value in the prose, so a full answer is stated.
- partial: some decisions are made in prose (e.g. which friends to drop, one house, the first moves), but not a full answer.
- none: no values of decision variables are committed in prose.

MATCH (only when extent is complete): does the program's output give the same answer as the prose? Answer "na" if the extent is not complete or the program printed no answer. Differences in formatting only (e.g. 9:00 vs 09:00) count as the same answer.

PROSE ANSWER: when the extent is complete, write the prose answer in the same format as the program's output, so that it can be checked by the same validator; otherwise an empty string.

Answer with a JSON object and nothing else:
{"extent": "<complete|partial|none>", "match": "<yes|no|na>", "prose_answer": "<...>"}

PROBLEM GIVEN TO THE MODEL:
{problem}

THE MODEL'S REASONING BEFORE THE PROGRAM{reasoning_note} (only the parts where it works with values or argues about solutions; "[...]" marks left-out parts such as code and restatements of the problem):
{reasoning}

THE PROGRAM'S OUTPUT:
{output}
\end{lstlisting}
\end{tcolorbox}
\end{listing*}

\begin{listing*}[p]
\caption{Prompt of the B judge, which labels the program B1--B4 with its evidence.}
\label{lst:judge-b}
\begin{tcolorbox}[colback=white, colframe=black, boxrule=0.5pt, width=\linewidth, arc=0pt, outer arc=0pt, left=4pt, right=4pt, top=3pt, bottom=3pt]
\begin{lstlisting}[style=prompt]
You are auditing a program written by a language model for a constraint problem. The model was asked to write a program (Python, Python with Z3, or PDDL for a planner) whose execution finds the answer. Before writing it, the model reasoned in prose, and that reasoning sometimes already solves the problem. Decide how much of the answer the program inherits from the prose instead of computing it.

LABEL (choose exactly one):
- hardcoded: the program prints an answer it does not compute. Examples: a literal itinerary, slot or grid is printed; Z3 is set up but its model is ignored; a PDDL domain is written so that only one action sequence is possible (e.g. step-counter predicates).
- partly_fixed: the program computes the answer, but some decisions were made in prose and written into the source. Examples: a fixed friend order with times recomputed along it; assignments asserted as `x == k` that no clue states; constraints deduced in prose added in place of the problem's own; friends or options ruled out in prose left out of the encoding; PDDL actions or preconditions written to impose a chosen order.
- steered: the program genuinely searches over the alternatives with the problem's constraints, but its tie-break, objective, preference reading or iteration order was chosen so that the search returns the answer the prose already found (e.g. "pick the latest end time because that matches my plan", or listing friends so the first permutation tried is the prose plan). Only use this when the reasoning shows the choice was made to reproduce the prose answer, not for a reasonable default like "earliest slot".
- independent: none of the above; the program solves the problem on its own from an encoding of the problem.

Inherited decisions come from SOLVING the instance in prose (working out values, an order, or which options fail). Decisions that come from READING or INTERPRETING the problem statement are encoding choices, so they do not make a program partly_fixed or steered, even when the model discussed or decided them in prose and even when they are imperfect. Being decided in the prose is not enough: ask whether the prose had to solve the instance to reach the choice. Examples that stay independent:
- a restriction read off the problem's own goal or constraints to approximate a constraint the formal language cannot express (e.g. only the goal's table blocks may be put on the table, standing in for "exactly 5 stacks");
- resolving a conflict between requirements the way the problem instructs (e.g. an additional constraint to end holding a block replaces that block's goal position);
- encoding a stated preference as a hard constraint or as the search order (e.g. "would rather not meet on Tuesday" excludes Tuesday);
- adding, dropping or changing actions, goals or constraints (an unfaithful encoding is a formalization error, not an inherited answer).
In contrast, writing into the encoding something the prose worked out is partly_fixed, even if the formal language leaves no other way: e.g. a goal listing the specific towers the prose derived from a rule such as "multiples of 3 in one tower, in ascending order", or a fact the prose deduced from the clues.

Answer with a JSON object and nothing else:
{"label": "<hardcoded|partly_fixed|steered|independent>", "evidence": "<one sentence naming the line or construct in the program that decided it>"}

PROBLEM GIVEN TO THE MODEL:
{problem}

THE MODEL'S REASONING BEFORE THE PROGRAM{reasoning_note}:
{reasoning}

THE PROGRAM:
{program}

ITS OUTPUT:
{output}
\end{lstlisting}
\end{tcolorbox}
\end{listing*}

\end{document}

%% file: tables/runs.tex
% generated by solver-like-behavior-in-llms/scripts/paper_runs_table.py; do not edit
\begin{tabular}{lllccc}
\toprule
 & & & \multicolumn{3}{c}{Not in FL} \\
\cmidrule(lr){4-6}
Model & Task & FL & 0 & A & B \\
\midrule
DeepSeek & Calendar & Python & 0 & 0 & 0 \\
 & Meeting & Python & 0 & 0 & 0 \\
 & ZebraLogic & Z3 & 0 & 0 & 1 \\
 & BlocksWorld & PDDL & -- & -- & -- \\
\midrule
GPT-OSS & Calendar & Python & 0 & 0 & 0 \\
 & Calendar & Z3 & 0 & 0 & 0 \\
 & Meeting & Python & 0 & 0 & 0 \\
 & Meeting & Z3 & 0 & 0 & 0 \\
 & ZebraLogic & Python & 0 & 0 & 0 \\
 & ZebraLogic & Z3 & 0 & 0 & 0 \\
 & BlocksWorld & Python & 0 & 0 & 0 \\
 & BlocksWorld & PDDL & -- & -- & -- \\
\midrule
Qwen & Calendar & Python & 0 & 0 & 0 \\
 & Calendar & Z3 & 0 & 0 & 0 \\
 & Meeting & Python & 0 & 0 & 0 \\
 & Meeting & Z3 & 3 & 5 & 2 \\
 & ZebraLogic & Python & 0 & 0 & 0 \\
 & ZebraLogic & Z3 & 0 & 0 & 0 \\
 & BlocksWorld & Python & 0 & 0 & 0 \\
 & BlocksWorld & PDDL & -- & -- & -- \\
\bottomrule
\end{tabular}

%% file: tables/ontology_cells.tex
% generated by solver-like-behavior-in-llms/scripts/paper_figures.py; do not edit
\begin{tabular}{llll|r|rrrrr|rrrrr|rr}
\toprule
Model & Task & FL & Cond. & Acc. & SLB & Cmpl. & Part. & Ratio & SLB tok. & B1 & B2 & B3 & B4 & n/a & C1 no & C2 no \\
\midrule
DeepSeek & Calendar & Python & 0 & 0.93 & 100 & 100 & 0 & 0.21 & 2{,}496 & 0 & 1 & 2 & 91 & 5 & 0/95 & 3/99 \\
 &  &  & A & 0.99 & 100 & 100 & 0 & 0.17 & 2{,}833 & 0 & 0 & 0 & 100 & 0 & 0/100 & 1/100 \\
 &  &  & B & 0.97 & 100 & 100 & 0 & 0.18 & 2{,}676 & 0 & 0 & 1 & 97 & 1 & 0/99 & 2/100 \\
\cmidrule(lr){2-17}
 & Meeting & Python & 0 & 0.92 & 98 & 97 & 1 & 0.28 & 15{,}561 & 0 & 2 & 13 & 76 & 7 & 7/89 & 0/90 \\
 &  &  & A & 0.93 & 100 & 90 & 10 & 0.22 & 13{,}529 & 0 & 0 & 6 & 88 & 6 & 11/83 & 2/87 \\
 &  &  & B & 0.93 & 100 & 98 & 2 & 0.24 & 14{,}284 & 0 & 0 & 9 & 81 & 6 & 8/90 & 0/90 \\
\cmidrule(lr){2-17}
 & ZebraLogic & Z3 & 0 & 0.94 & 100 & 98 & 2 & 0.22 & 4{,}189 & 0 & 0 & 0 & 94 & 6 & 0/92 & 0/97 \\
 &  &  & A & 0.94 & 99 & 85 & 14 & 0.16 & 3{,}401 & 0 & 0 & 0 & 98 & 2 & 0/84 & 3/85 \\
 &  &  & B & 0.96 & 100 & 95 & 5 & 0.21 & 3{,}833 & 0 & 1 & 0 & 97 & 2 & 0/93 & 1/95 \\
\cmidrule(lr){2-17}
 & BlocksWorld & PDDL & 0 & 0.76 & 84 & 38 & 46 & 0.06 & 2{,}245 & 0 & 5 & 0 & 94 & 0 & 18/28 & 4/27 \\
 &  &  & A & 0.76 & 88 & 34 & 54 & 0.05 & 1{,}882 & 0 & 5 & 0 & 95 & 0 & 19/30 & 3/29 \\
 &  &  & B & 0.71 & 67 & 28 & 39 & 0.04 & 1{,}538 & 0 & 5 & 0 & 94 & 1 & 11/18 & 5/17 \\
\midrule
GPT-OSS & Calendar & Python & 0 & 0.98 & 99 & 98 & 1 & 0.30 & 476 & 0 & 0 & 0 & 99 & 1 & 2/98 & 2/98 \\
 &  &  & A & 0.97 & 84 & 81 & 3 & 0.23 & 369 & 0 & 0 & 0 & 98 & 2 & 3/80 & 3/81 \\
 &  &  & B & 0.97 & 66 & 65 & 1 & 0.16 & 256 & 0 & 0 & 1 & 97 & 2 & 1/64 & 0/65 \\
 &  & Z3 & 0 & 0.96 & 88 & 88 & 0 & 0.22 & 355 & 0 & 0 & 0 & 97 & 3 & 5/85 & 2/86 \\
 &  &  & A & 0.91 & 85 & 84 & 1 & 0.21 & 344 & 0 & 2 & 0 & 91 & 7 & 1/76 & 1/80 \\
 &  &  & B & 0.95 & 64 & 63 & 1 & 0.15 & 220 & 0 & 1 & 0 & 95 & 4 & 2/61 & 2/61 \\
\cmidrule(lr){2-17}
 & Meeting & Python & 0 & 0.66 & 98 & 85 & 13 & 0.45 & 1{,}732 & 40 & 11 & 0 & 42 & 7 & 19/72 & 25/68 \\
 &  &  & A & 0.74 & 78 & 60 & 18 & 0.26 & 692 & 0 & 15 & 2 & 76 & 7 & 14/40 & 6/35 \\
 &  &  & B & 0.79 & 68 & 56 & 12 & 0.23 & 519 & 1 & 7 & 0 & 81 & 10 & 22/48 & 10/41 \\
 &  & Z3 & 0 & 0.38 & 69 & 63 & 6 & 0.19 & 724 & 0 & 14 & 0 & 60 & 25 & 9/29 & 6/23 \\
 &  &  & A & 0.44 & 62 & 46 & 16 & 0.13 & 354 & 0 & 13 & 0 & 66 & 21 & 4/26 & 2/22 \\
 &  &  & B & 0.33 & 58 & 41 & 17 & 0.12 & 317 & 0 & 9 & 0 & 58 & 33 & 8/21 & 4/18 \\
\cmidrule(lr){2-17}
 & ZebraLogic & Python & 0 & 0.86 & 78 & 56 & 22 & 0.24 & 548 & 0 & 17 & 0 & 75 & 8 & 3/55 & 4/56 \\
 &  &  & A & 0.86 & 52 & 29 & 23 & 0.12 & 174 & 0 & 11 & 0 & 80 & 9 & 0/29 & 0/29 \\
 &  &  & B & 0.85 & 66 & 40 & 26 & 0.18 & 365 & 0 & 11 & 0 & 80 & 9 & 2/40 & 1/40 \\
 &  & Z3 & 0 & 0.84 & 60 & 47 & 13 & 0.16 & 345 & 0 & 2 & 0 & 86 & 12 & 1/45 & 2/47 \\
 &  &  & A & 0.87 & 68 & 49 & 19 & 0.17 & 364 & 0 & 5 & 0 & 87 & 8 & 0/47 & 1/49 \\
 &  &  & B & 0.85 & 37 & 28 & 9 & 0.07 & 124 & 0 & 1 & 0 & 95 & 4 & 2/27 & 2/28 \\
\cmidrule(lr){2-17}
 & BlocksWorld & Python & 0 & 0.34 & 56 & 40 & 16 & 0.18 & 399 & 7 & 6 & 0 & 85 & 2 & 12/36 & 13/36 \\
 &  &  & A & 0.28 & 35 & 20 & 15 & 0.09 & 191 & 1 & 4 & 0 & 95 & 0 & 5/16 & 6/16 \\
 &  &  & B & 0.34 & 37 & 27 & 10 & 0.11 & 185 & 5 & 2 & 0 & 93 & 0 & 9/26 & 13/26 \\
 &  & PDDL & 0 & 0.52 & 7 & 2 & 5 & 0.01 & 22 & 0 & 4 & 0 & 96 & 0 & 1/1 & 0/0 \\
 &  &  & A & 0.54 & 10 & 3 & 7 & 0.02 & 49 & 0 & 5 & 0 & 95 & 0 & 0/0 & 0/0 \\
 &  &  & B & 0.45 & 7 & 1 & 6 & 0.01 & 24 & 0 & 4 & 0 & 96 & 0 & 0/0 & 0/0 \\
\midrule
Qwen & Calendar & Python & 0 & 0.97 & 100 & 100 & 0 & 0.21 & 1{,}291 & 0 & 0 & 3 & 96 & 1 & 3/99 & 3/100 \\
 &  &  & A & 0.97 & 100 & 100 & 0 & 0.15 & 1{,}174 & 0 & 0 & 1 & 98 & 1 & 1/99 & 3/100 \\
 &  &  & B & 0.96 & 100 & 100 & 0 & 0.13 & 1{,}178 & 0 & 0 & 1 & 98 & 1 & 3/99 & 4/100 \\
 &  & Z3 & 0 & 0.85 & 100 & 100 & 0 & 0.17 & 1{,}154 & 0 & 0 & 1 & 86 & 13 & 0/87 & 2/97 \\
 &  &  & A & 0.88 & 99 & 99 & 0 & 0.14 & 1{,}161 & 0 & 0 & 3 & 87 & 10 & 1/89 & 1/91 \\
 &  &  & B & 0.81 & 100 & 100 & 0 & 0.12 & 1{,}076 & 0 & 0 & 0 & 82 & 18 & 0/81 & 1/93 \\
\cmidrule(lr){2-17}
 & Meeting & Python & 0 & 0.89 & 100 & 100 & 0 & 0.31 & 10{,}434 & 0 & 0 & 9 & 85 & 6 & 27/92 & 11/92 \\
 &  &  & A & 0.88 & 100 & 90 & 10 & 0.25 & 7{,}571 & 0 & 0 & 3 & 91 & 6 & 21/83 & 7/80 \\
 &  &  & B & 0.96 & 100 & 98 & 2 & 0.29 & 10{,}079 & 0 & 2 & 5 & 91 & 2 & 23/92 & 11/88 \\
 &  & Z3 & 0 & 0.27 & 100 & 94 & 6 & 0.17 & 6{,}397 & 0 & 1 & 0 & 28 & 71 & 8/23 & 11/65 \\
 &  &  & A & 0.25 & 96 & 79 & 17 & 0.11 & 3{,}611 & 0 & 0 & 1 & 25 & 74 & 6/20 & 8/50 \\
 &  &  & B & 0.18 & 100 & 83 & 17 & 0.14 & 4{,}876 & 0 & 0 & 0 & 23 & 77 & 8/18 & 11/48 \\
\cmidrule(lr){2-17}
 & ZebraLogic & Python & 0 & 0.95 & 100 & 100 & 0 & 0.20 & 2{,}360 & 0 & 5 & 0 & 91 & 4 & 4/96 & 4/100 \\
 &  &  & A & 0.95 & 98 & 87 & 11 & 0.16 & 1{,}520 & 0 & 0 & 0 & 96 & 4 & 2/85 & 4/87 \\
 &  &  & B & 0.96 & 97 & 95 & 2 & 0.14 & 2{,}312 & 0 & 2 & 0 & 94 & 4 & 1/90 & 1/93 \\
 &  & Z3 & 0 & 0.59 & 100 & 100 & 0 & 0.22 & 2{,}014 & 0 & 2 & 0 & 60 & 38 & 3/62 & 4/95 \\
 &  &  & A & 0.59 & 97 & 78 & 19 & 0.13 & 1{,}011 & 0 & 0 & 0 & 59 & 41 & 1/46 & 2/65 \\
 &  &  & B & 0.56 & 94 & 91 & 3 & 0.15 & 1{,}322 & 0 & 2 & 0 & 54 & 44 & 1/54 & 4/78 \\
\cmidrule(lr){2-17}
 & BlocksWorld & Python & 0 & 0.66 & 100 & 100 & 0 & 0.20 & 5{,}742 & 0 & 8 & 0 & 91 & 0 & 21/100 & 29/99 \\
 &  &  & A & 0.67 & 100 & 96 & 4 & 0.16 & 4{,}551 & 0 & 4 & 1 & 95 & 0 & 24/96 & 28/95 \\
 &  &  & B & 0.67 & 100 & 99 & 1 & 0.16 & 4{,}577 & 0 & 8 & 0 & 92 & 0 & 28/99 & 28/98 \\
 &  & PDDL & 0 & 0.71 & 62 & 41 & 21 & 0.04 & 913 & 0 & 10 & 0 & 90 & 0 & 21/30 & 9/29 \\
 &  &  & A & 0.69 & 63 & 21 & 42 & 0.04 & 835 & 0 & 7 & 0 & 93 & 0 & 9/13 & 4/12 \\
 &  &  & B & 0.69 & 63 & 27 & 36 & 0.03 & 765 & 0 & 7 & 0 & 93 & 0 & 16/21 & 7/19 \\
\bottomrule
\end{tabular}